\documentclass[sigconf]{acmart}
\copyrightyear{2021}
\acmYear{2021}
\setcopyright{acmlicensed}
\acmConference[SIGIR '21]{Proceedings of the 44th International ACM SIGIR Conference on Research and Development in Information Retrieval}{July 11--15, 2021}{Virtual Event, Canada}
\acmBooktitle{Proceedings of the 44th International ACM SIGIR Conference on Research and Development in Information Retrieval (SIGIR '21), July 11--15, 2021, Virtual Event, Canada}
\acmPrice{15.00}
\acmDOI{10.1145/3404835.3462874}
\acmISBN{978-1-4503-8037-9/21/07}

\usepackage{url}
\usepackage{hyperref}
\usepackage{booktabs}
\usepackage{graphicx}
\usepackage{subfigure}
\usepackage{bm}
\usepackage{color}
\usepackage{amsmath}
\usepackage{amsfonts}
\usepackage{stmaryrd}
\usepackage{stfloats}
\usepackage{xspace}
\usepackage{multirow}
\usepackage{pifont}
\usepackage{pbox}

\newcommand{\tabincell}[2]{\begin{tabular}{@{}#1@{}}#2\end{tabular}}
\newcommand{\ie}{\emph{i.e.,}\xspace}
\newcommand{\eg}{\emph{e.g.,}\xspace}

\makeatletter
\def\@fnsymbol#1{\ensuremath{\ifcase#1\or \dagger\or \ddagger\or
   \mathsection\or \mathparagraph\or \|\or **\or \dagger\dagger
   \or \ddagger\ddagger \else\@ctrerr\fi}}
\makeatother

\begin{document}
\fancyhead{}

%% The "title" command has an optional parameter, allowing the author to define a "short title" to be used in page headers.
\title{Video Corpus Moment Retrieval with Contrastive Learning}

%% The "author" command and its associated commands are used to define the authors and their affiliations. Of note is the shared affiliation of the first two authors, and the "authornote" and "authornotemark" commands used to denote shared contribution to the research.
\author{Hao Zhang, Aixin Sun}
\authornote{Aixin Sun is the corresponding author.}
\affiliation{\institution{Nanyang Technological University, Singapore}\country{}}
\email{{hao007@e., axsun@}ntu.edu.sg}

\author{Wei Jing}
\affiliation{\institution{Institute of Infocomm Research, A*STAR, Singapore}\country{}}
\email{21wjing@gmail.com}

\author{Guoshun Nan}
\affiliation{
\institution{Singapore University of Technology and Design, Singapore}\country{}}
\email{nanguoshun@gmail.com}

\author{Liangli Zhen}
\affiliation{\institution{Institute of High Performance Computing, A*STAR, Singapore}\country{}}
\email{zhenll@ihpc.a-star.edu.sg}

\author{Joey Tianyi Zhou}
\affiliation{\institution{Institute of High Performance Computing, A*STAR, Singapore}\country{}}
\email{zhouty@ihpc.a-star.edu.sg}

\author{Rick Siow Mong Goh}
\affiliation{\institution{Institute of High Performance Computing, A*STAR, Singapore}\country{}}
\email{gohsm@ihpc.a-star.edu.sg}

%% other information printed in the page headers. This command allows the author to define a more concise list of authors' names for this purpose.
\renewcommand{\shortauthors}{Zhang and Sun, et al.}

\begin{abstract}
Given a collection of untrimmed and unsegmented videos, \textit{video corpus moment retrieval} (VCMR) is to retrieve a temporal moment (\ie a fraction of a video) that semantically corresponds to a given text query. As video and text are from two distinct feature spaces, there are two general approaches to address VCMR: (i) to separately encode each modality representations, then align the two modality representations for query processing, and (ii) to adopt fine-grained cross-modal interaction to learn multi-modal representations for query processing. While the second approach often leads to better retrieval accuracy, the first approach is far more efficient. In this paper, we propose a \textbf{Re}trieval and \textbf{Lo}calization \textbf{Net}work with \textbf{C}ontrastive \textbf{L}earning (ReLoCLNet) for VCMR. We adopt the first approach and introduce two contrastive learning objectives to refine video encoder and text encoder to learn video and text representations separately but with better alignment for VCMR. The video contrastive learning (VideoCL) is to maximize mutual information between query and candidate video at video-level. The frame contrastive learning (FrameCL) aims to highlight the moment region corresponds to the query at frame-level, within a video. Experimental results show that, although ReLoCLNet encodes text and video separately for efficiency, its retrieval accuracy is comparable with baselines adopting cross-modal interaction learning.\footnote{\url{https://github.com/IsaacChanghau/ReLoCLNet}}    
\end{abstract}




\maketitle

% =============================================
\section{Introduction}\label{sec:intro}
% =============================================

\begin{figure}[t]
    \centering
    \includegraphics[trim={0cm 0cm 0cm 0cm},clip,width=\linewidth]{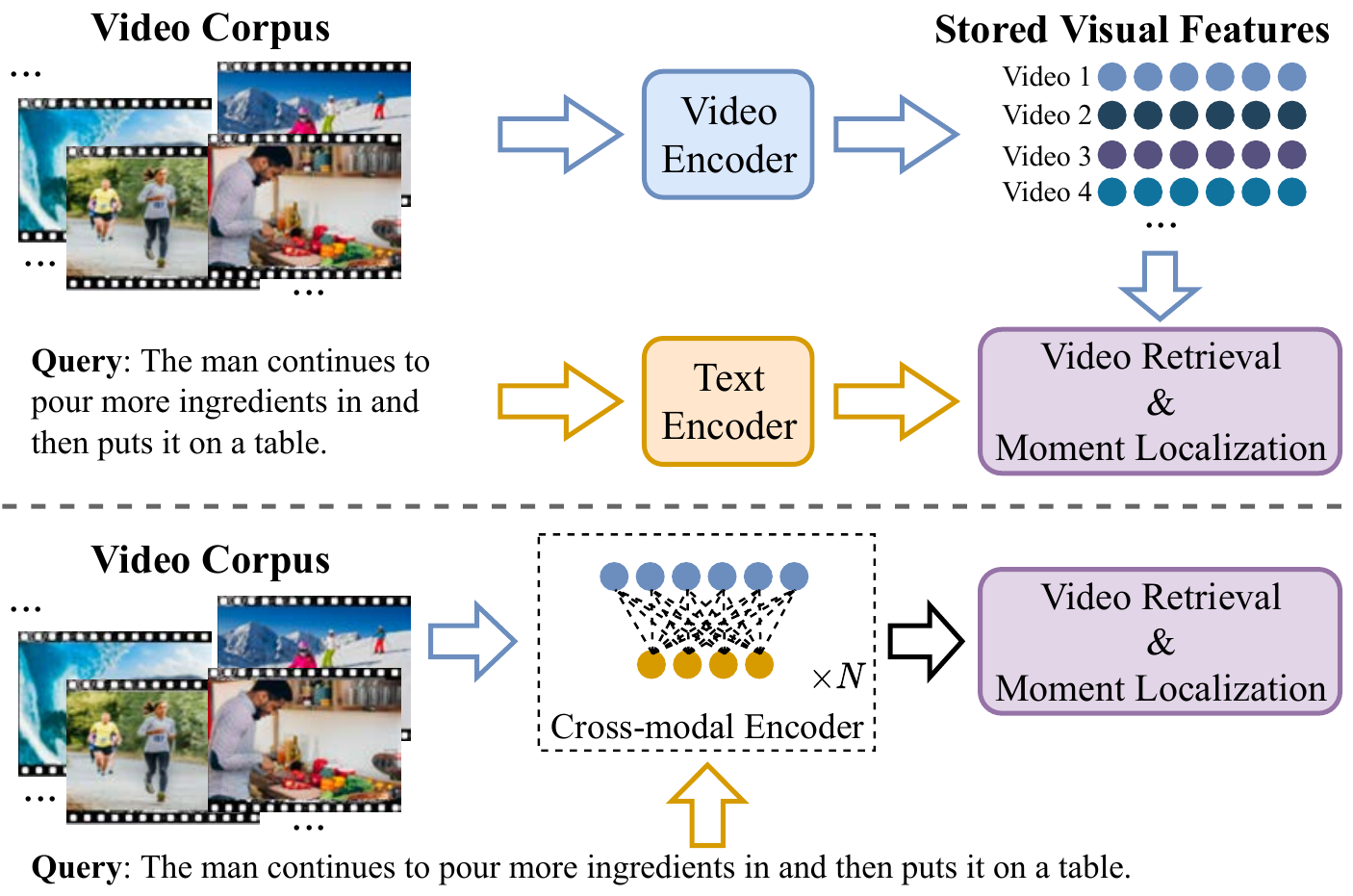}
    \caption{Two approaches to VCMR: unimodal encoding vs. cross-modal interaction learning.}
	\label{fig:interaction_example}
\end{figure}

Video corpus moment retrieval (VCMR) is a domain-specific retrieval task. The aim is to retrieve a short fraction in a video, that  semantically corresponds to a text query, from a corpus of untrimmed and unsegmented videos. We use the term \textit{VCMR} to distinguish this task from \textit{single video moment retrieval} (SVMR). As its name suggests, SVMR is to retrieve a short fraction from a single given video that corresponds to a text query. In fact, the task of VCMR was extended from SVMR by \citet{escorcia2019temporal}. VCMR better matches real-world application scenarios, such as query-based video surveillance, search, and navigation, within a video corpus. 

Video and text are from two feature spaces. In order to perform query-based video moment retrieval, we need to learn the matching between query and video from training samples.  In general, there are two approaches, illustrated in Figure~\ref{fig:interaction_example}. One is to encode video and text separately, and learn the matching through late feature fusion, known as \textbf{\textit{unimodal encoding}}~\cite{escorcia2019temporal,lei2020tvr}. With unimodal encoding, the text query is encoded to a $d$-dimensional feature vector. A video is encoded to a sequence of $d$-dimensional feature vectors, where each vector corresponds to a small fraction of the video, \eg a few frames. The other is \textbf{\textit{cross-modal interaction learning}}, which takes in a video as a sequence of visual features, and the query as a sequence of word features to learn their interactions~\cite{zhang2020hierarchical}. The latter typically leads to better retrieval accuracy as the learned parameters capture the relevance between query and video at fine-grained granularity. However, in query evaluation, cross-modal encoding needs to be performed between query and \textit{every video} in corpus (illustrated by ``$\times N$'' in Figure~\ref{fig:interaction_example}), leading to high computational cost. On the other hand, with unimodal encoding, visual features can be pre-encoded and stored. In query evaluation, we only need to encode query and then perform video retrieval and moment localization. The challenge becomes to refine two separate encoders during training process, such that the encoded features are well aligned for accurate retrieval. We follow the unimodal encoding approach for its high efficiency. As illustrated in Figure~\ref{fig:interaction_example}, for both approaches, video retrieval and moment localization are performed jointly, \ie the model is trained with a joint objective. An earlier study~\cite{lei2020tvr} has shown that joint learning outperforms two-stage learning where video retrieval and moment localization are treated as two separate subtasks and performed in stages. 

The essence of cross-modal interaction is to highlight the relevant and important information from both modalities via co-attention mechanisms. Meanwhile, contrastive learning~\cite{hadsell2006dimensionality,wu2018unsupervised,he2020momentum} is a strategy to maximize the mutual information (MI)~\cite{bell1995information,hyvarinen2000independent} of positive pairs and to minimize the MI of negative pairs. In our context, a pair of matching video and query is a positive pair and a non-matching pair is a negative pair in training. We consider that both cross-modal interaction learning and contrastive learning share the same objective of emphasizing the relevant information of input pairs. Hence, we can apply contrastive learning to refine encoders in unimodal encoding to achieve similar effectiveness.  

In this paper, we develop a Retrieval and Localization Network (ReLoNet) as a base network to separately encode video and query representations, and to (late) fuse them for joint retrieval. We then introduce contrastive learning to ReLoNet to simulate cross-modal interactions between video and query, and propose ReLoCLNet. Build on top of ReLoNet, ReLoCLNet is trained with two contrastive learning objectives: VideoCL and FrameCL. The VideoCL objective aims to learn video and text features such that the semantically related videos and queries are close to each other, and far away otherwise. The FrameCL works at frame-level for moment localization, which simulates fine-grained cross-modal interactions between visual and textual features within a video. In FrameCL, we regard the features within target moment as foreground (positive samples), while the remaining as background (negative samples). Thus, FrameCL enhances the MI between query and foreground, while suppresses the MI between query and background. Once trained, the learned parameters in video encoder and text encoder can be used to encode video and text features separately and independently. Accordingly, all videos in a given corpus can be pre-encoded by the learned video encoder and stored, as illustrated in Figure~\ref{fig:interaction_example}, for efficient retrieval. Our main contributions are as follows:
\begin{itemize}
    \item To the best of our knowledge, we are the first to address the contradiction between high efficiency and high-quality retrieval in VCMR, by replacing conventional cross-modal interaction learning with contrastive learning.
    \item We propose two contrastive learning objectives, VideoCL and FrameCL, to simulate cross-modal interactions at both video level and frame level, by measuring the mutual information between video and query at different granularity.
    \item We conduct experiments on two benchmarks to demonstrate that ReLoCLNet achieves comparable accuracy with cross-modal interaction learning, with much faster retrieval speed. On TVR dataset,  ReLoCLNet is about 56 times faster. 
\end{itemize}

% =============================================
\section{Related Work}\label{sec:related}
% =============================================
We review related studies on video retrieval, single video moment retrieval, video corpus moment retrieval, and contrastive learning.

\paragraph{Video Retrieval.} Given a text query and a set of candidate videos, video retrieval (VR) aims to retrieve and rank  candidate videos based on their relevance to the query. Many  works~\cite{xu2015jointly,pan2016jointly,mithun2018learning} jointly model video and text to map them into two holistic representations in a joint embedding space. Their similarities are computed as ranking scores. \citet{venugopalan2015sequence} develop a sequence to sequence model for video to text translation and matching. To handle long text query, hierarchical models~\cite{shao2018find,zhang2018cross} are proposed to match video and text at different scales. Recently, \citet{li2020sea} present a text-video matching strategy by using multiple encoders, which can prevent matching from being dominated by a specific encoder. 

\paragraph{Single Video Moment Retrieval} SVMR aims to localize a relevant temporal moment in an untrimmed video for a given query~\cite{Hendricks2017LocalizingMI,Gao2017TALLTA}. This problem has been well studied and many approaches have been proposed. Ranking-based methods~\cite{hendricks2017localizing,Gao2017TALLTA,hendricks2018localizingM,Liu2018CML,chen2019semantic,ge2019mac,zhang2019exploiting} solve SVMR with propose-and-rank pipeline. The given video is pre-segmented into proposals (\ie video segments) and the proposals are ranked by their similarities to the query.  Anchor-based methods~\cite{chen2018temporally,yuan2019semantic,zhu2019cross,Wang2020TemporallyGL} replace proposal generation process by assigning each frame with multi-scale anchors sequentially. The anchor (similar to temporal window on video) with highest confidence is selected as result. Regression-based methods~\cite{lu2019debug,Yuan2019ToFW,chen2020rethinking,chen2020learning,mun2020local,zeng2020dense} directly regress temporal times of target moment through cross-modal interactions learning. Span-based methods~\cite{ghosh2019excl,rodriguez2020proposal,zhang2020vslnet,zhang2021qa4nlvl} follow the concept of extractive question answering (QA)~\cite{Seo2017BidirectionalAF,wang2017gated,huang2018fusionnet,wei2018fast}. These methods adopt QA based models to encode multimodal representations for video and query, and predict start and end boundaries of target moment directly. There are also studies~\cite{Wang2019LanguageDrivenTA,he2019Readwa,cao2020strong,hahn2020tripping,wu2020reinforcement,Wu2020TreeStructuredPB} formulate SVMR as sequential decision-making problem and design reinforcement learning methods. Other solutions such as weakly supervised learning and jointly training with event captioning have also been explored~\cite{mithun2019weakly,Lin2020WeaklySupervisedVM,wu2020reinforcement,shao2018find,zhang2020learning,chen2020learning,wang2020dual}.

\paragraph{Video Corpus Moment Retrieval} \citet{escorcia2019temporal} first extend SVMR to VCMR, and devise a ranking-based clip-query alignment model. The model compares query features with uniformly partitioned video clips. \citet{lei2020tvr} construct a new VCMR dataset named TVR, where the videos come with textual subtitles. The authors propose a proposal-free cross-modal moment localization (XML) model to jointly learn video retrieval and moment localization objectives. Note that, the ``cross-modal'' component in XML conceptually is the same as late feature fusion in unimodal encoding approach (see Figure~\ref{fig:interaction_example}). In our classification, the XML model falls under unimodal encoding approach. As a typical cross-modal interaction learning approach,  \citet{zhang2020hierarchical} propose a hierarchical multimodal encoder (HAMMER) to jointly train video retrieval and moment localization with fine-grained cross-modal interactions between query and video. Though effective, HAMMER suffers from low-efficiency and high computational cost. Lastly, \citet{li2020hero} develops a video-language model for joint representation learning. The model is applied on VCMR for fine-tuning purpose only.

\paragraph{Contrastive learning} Contrastive learning (CL) usually serves as an unsupervised objective to learn representations by contrasting positive pairs against negative pairs~\cite{hadsell2006dimensionality,wu2018unsupervised,zhuang2019local,chen2020a,misra2020self,he2020momentum}. In our context, a positive pair is a matching video-query pair, and a negative pair is a non-matching video and query. One way to achieve contrastive learning is to directly maximize the mutual information (MI)~\cite{bell1995information,hyvarinen2000independent} between latent representations~\cite{hjelm2018learning,philip2019learning}. There are also solutions to estimate the lower bounds of MI~\cite{belghazi2018mine,oord2018representation} for unsupervised learning~\cite{velickovic2018deep,hjelm2018learning,Sun2020InfoGraph}. CL has been applied to vision-language tasks to learn the joint representations of visual and textual modalities~\cite{sun2019cbt,luo2020univilm,miech2020end}.  \citet{miech2020end} proposes a MIL-NCE objective to address the misalignment between text and video clip in narrated videos for joint representation learning. While MIL-NCE is used for video-text matching, \citet{luo2020univilm} develops a unified pre-training model for multimodal understanding and generation.

% =============================================
\section{The R\lowercase{e}L\lowercase{o}CLN\lowercase{et} Model}\label{sec:method}
% =============================================
To ensure retrieval efficiency, we follow the unimodal encoding approach (see Figure~\ref{fig:interaction_example}) and aim to develop video encoder and text encoder for effective feature encoding separately. To achieve high-quality retrieval results, we aim to simulate the cross-modal interaction learning to better align the encoded video and text features. To this end, we introduce contrastive learning to our model. Conceptually, contrastive learning and cross-modal interaction learning share a similar objective of highlighting the relevant information of input pairs, \ie matching video-query pairs in our setting. Different from cross-modal interaction learning,  contrastive learning is only engaged in the training phase. Once trained, the learned parameters ensure the alignment between the encoded video features and text features even though the two features are encoded separately.  The task objective, \ie video retrieval and moment localization, can then be easily achieved through late feature fusion. 

\begin{figure*}[t]
    \centering
    \includegraphics[trim={0cm 0cm 2cm 0cm},clip,width=0.8\textwidth]{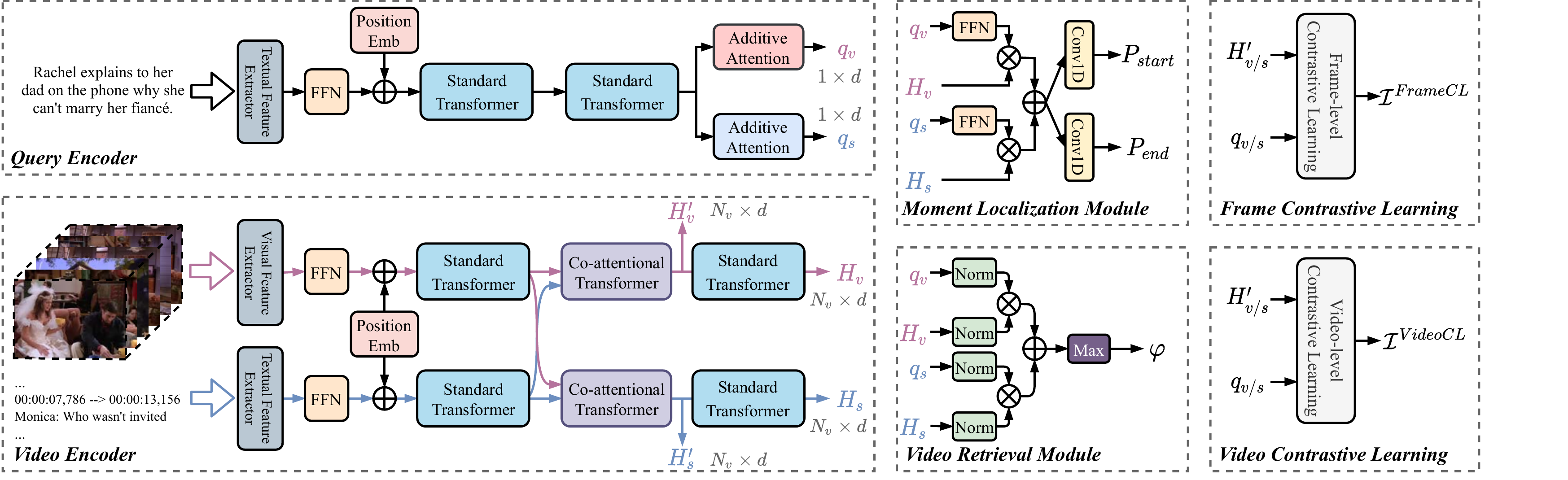}
    \caption{In both ReLoNet and ReLoCLNet, query is encoded to $\bm{q}_{v/s}$ and video is encoded to $\bm{H}_{v/s}$, for video retrieval and moment localization. ReLoCLNet adds contrastive learning objectives through $\bm{q}_{v/s}$ and $\bm{H'}_{v/s}$ to refine query and video encoders.}
	\label{fig:architecture}
\end{figure*}

In this section, we first develop the ReLoNet as a base model, to separately encode video and query inputs and fuse them for prediction. Then we design two contrastive learning objectives: (i) Video-level Contrastive Learning (VideoCL) for video retrieval, and (ii) Frame-level Contrastive Learning (FrameCL) for moment localization. During training phrase, VideoCL and FrameCL simulate the cross-modal interaction to enhance the representation learning. During inference (\ie retrieval) phrase, the model separately encodes video and query to maintain retrieval efficiency. The overall architecture of the proposed model is shown in Figure~\ref{fig:architecture}. Next, we formally formulate the research problem,  then detail the components in ReLoNet and ReLoCLNet.

% =============================================
\subsection{Problem Formulation}\label{ssec:formulation}
% =============================================
We denote a video corpus as $\mathcal{V}=\{V^1,V^2,\dots,V^M\}$, where $M$ is the number of videos and $V^k=[f_i]_{i=0}^{T-1}$ represents the $k$-th video with $T$ frames.\footnote{Videos in $\mathcal{V}$ could be of  different lengths; we simply use $T$ to represent the length (in number of frames) of an arbitrary video.} Given a text query $Q=[q_i]_{i=0}^{n_q-1}$, we aim to retrieve the temporal moment (starting from $\tau^s$ and ending at $\tau^e$) in $V^*$ that semantically corresponds to $Q$ from video corpus $\mathcal{V}$. Here $V^*$ denotes a video that contains the ground truth moment and $\tau^{s/e}$ are the start/end time points of target moment in $V^*$. Thus, VCMR has two objectives: (i) video retrieval, \ie finding $V^*$ from $\mathcal{V}$; and (ii) moment localization, \ie locating the target moment in $V^*$.

For words in $Q$, the initial encoding is obtained from pre-trained word embeddings or language models as $\bm{Q}=[\bm{q}_i]_{i=0}^{n_q-1}\in\mathbb{R}^{d_w\times n_q}$, where $d_w$ is the word feature dimension. For each video $V\in \mathcal{V}$, we split it into $n_v$ clip units, and use pre-trained feature extractor to encode them into visual features $\bm{V}=[\bm{v}_i]_{i=0}^{n_v-1}\in\mathbb{R}^{d_v\times n_v}$, where $d_v$ is the visual feature dimension. Then, $\tau^{s(e)}$ are mapped to the corresponding indices $i^{s(e)}$ in the visual feature sequence, and the target moment is represented as $\bm{m}^*=\{\bm{v}_i|i=i^s,\cdots,i^e\}$, where $0\leq i^s \leq i^e \leq n_v-1$. That is, in term of visual feature space, $\bm{m}^*$ may correspond to a sequence of $\bm{v}_i$'s of any length within $n_v$ starting from any index. The best matching $\bm{m}^*$ can be estimated by: 
\begin{equation}
    \bm{m}^* = \arg\max_{\bm{m}\in\bm{V}, \bm{V} \in \mathcal{V}} p(\bm{m}|\bm{V},\bm{Q})p(\bm{V}|\bm{Q})
\label{eq:approx_objective}
\end{equation}
Given $M$ videos in $\mathcal{V}$ with average video feature length $n_v$, the search space is $\mathcal{O}(M\times n_v^2)$. It is infeasible to compute $\bm{m}^*$ in such a large space. Hence, we approximate Eq.~\ref{eq:approx_objective} by:
\begin{equation}
    \bm{V}^*=\arg\max_{\bm{V}}p(\bm{V}|\bm{Q})\;\;\text{and}\;\;\bm{m}^*\approx\arg\max_{\bm{m}\in\bm{V}^*}p(\bm{m}|\bm{V}^*,\bm{Q})
\label{eq:final_objective}
\end{equation}
Eq.~\ref{eq:final_objective} is consistent with two objectives of VCMR, \eg video retrieval and moment localization. The search space reduces to $\mathcal{O}(M + M'\times n_v^2)$, where $M'$ is the top-$M'$ retrieved videos ($M'\ll M$) from the video corpus. 
In addition to visual features, a video may contain its own multi-modality features, such as subtitle and audio. For instance, videos in TVR dataset~\cite{lei2020tvr} come with subtitles. We denote the subtitle of a video by $S$, and the features extracted from subtitle by $\bm{S}\in\mathbb{R}^{d_w\times n_v}$. For easy presentation, we assume all videos come with subtitles and simply use ``video'' to refer ``video + subtitle''.

% =============================================
\subsection{Query Encoder}\label{ssec:query_encoder}
% =============================================
The structure of query encoder is shown in Figure~\ref{fig:architecture}. Given a text query $Q$ with $n_q$ words, we first apply textual feature extractor to covert words in the query to corresponding features $\bm{Q}=[\bm{q}_i]_{i=0}^{n_q-1}\in\mathbb{R}^{d_w\times n_q}$. Then we project the obtained features into dimension $d$ with a feed-forward layer as $\bm{\hat{Q}}=\bm{W}_{q}\cdot\bm{Q}+\bm{b}_{q}\in\mathbb{R}^{d\times n_q}$, where $\bm{W}_{q}\in\mathbb{R}^{d\times d_w}$ and $\bm{b}_q\in\mathbb{R}^{d}$ are the learnable weight and bias, respectively.

Positional embedding is incorporated to each feature of the query sequence $\bm{\hat{Q}}$ before they are fed to the transformer blocks~\cite{vaswani2017attention}.  We adopt the transformer block to better capture the contextual representations of the query, for its proven effectiveness~\cite{lu2019debug,chen2020rethinking,zhang2020vslnet}. Specifically, the transformer block consists of a multi-head attention layer and a feed-forward layer. Residual connection~\cite{he2016resnet} and layer normalization~\cite{Ba2016LayerN} strategies are applied to each layer in the transformer block. The encoded contextual representations of the query after the transformer block become $\bm{\widetilde{Q}}$.
\begin{equation}
    \bm{\widetilde{Q}} = \mathtt{Transformer}_{q}(\bm{\hat{Q}})
\label{eq:self_transformer}
\end{equation}
We use \textit{two} transformer blocks in the query encoder. Then we apply additive attention mechanism~\cite{Bahdanau2015NeuralMT} to compute the attention scores of each query word. The scores computed are utilized to aggregate the information of  $\bm{\widetilde{Q}}=[\bm{\tilde{q}}_0,\bm{\tilde{q}}_2,\dots,\bm{\tilde{q}}_{n_q-1}]$ to compute the modularized query vector, \ie the sentence representation of $\bm{\widetilde{Q}}$:
\begin{equation}
    \bm{\alpha}^q = \mathtt{Softmax}\big(\bm{W}_{m,\alpha}\cdot\bm{\widetilde{Q}})\big) \in \mathbb{R}^{n_q},\;\;
    \bm{q}_{m} = \sum_{i=0}^{n_q-1}\alpha_i^q\times\bm{q}_i \in \mathbb{R}^{d}
\end{equation}
where $\bm{q}_{m}\in\mathbb{R}^{d}$ denotes the modularized query vector. $m\in\{v,s\}$ means two modularized query vectors, $\bm{q}_{v}$ and $\bm{q}_{s}$, are computed for matching with visual and subtitle features, respectively. Both $\bm{q}_{v}$ and $\bm{q}_{s}$ are $d$-dimensional vectors as shown in Figure~\ref{fig:architecture}. If the videos to be retrieved do not contain subtitles, then only $\bm{q}_{v}$ is computed. 

% =============================================
\subsection{Video Encoder}\label{ssec:video_encoder}
% =============================================
We detail the video encoder with the assumption that the videos come with subtitles, as shown in Figure~\ref{fig:architecture}. Given a video with its subtitle, we first use visual and textual feature extractors to obtain the corresponding visual and subtitle features $\bm{V}\in\mathbb{R}^{d_v\times n_v}$ and $\bm{S}\in\mathbb{R}^{d_w\times n_v}$, respectively. Then both $\bm{V}$ and $\bm{S}$ are projected into dimension $d$ with two feed-forward layers as $\bm{\hat{V}} =\bm{W}_{v}\cdot\bm{V}+\bm{b}_{v}\in\mathbb{R}^{d\times n_v}$ and $\bm{\hat{S}} =\bm{W}_{s}\cdot\bm{S}+\bm{b}_{s}\in\mathbb{R}^{d\times n_v}$, where $\bm{W}_{v}\in\mathbb{R}^{d\times d_v}$ and $\bm{b}_v\in\mathbb{R}^{d}$ are the weight and bias for video feed-forward layer; $\bm{W}_{s}\in\mathbb{R}^{d\times d_w}$ and $\bm{b}_s\in\mathbb{R}^{d}$ are the weight and bias for subtitle feed-forward layer. 

Similar to the query encoder, we add positional embeddings to both $\bm{\hat{V}}$ and $\bm{\hat{S}}$, and feed them to the transformer block. The encoded contextual representations for video and subtitle are:
\begin{equation}
    \bm{\widetilde{V}} =\mathtt{Transformer}_{v}(\bm{\hat{V}}),\;\;
    \bm{\widetilde{S}} =\mathtt{Transformer}_{s}(\bm{\hat{S}})
\end{equation}
where $\bm{\widetilde{V}}\in\mathbb{R}^{d\times n_v}$ and $\bm{\widetilde{S}}\in\mathbb{R}^{d\times n_v}$.

Different from the query encoder, we do not use two transformer blocks here. Instead, after the first transformer blocks, we use co-attentional transformer blocks~\cite{tan2019lxmert,lu2019vilbert,zhu2020actbert,lei2020tvr}. Because the visual content in a video and its subtitle are well aligned, through co-attentional transformers, we are able to better capture the cross-modal representations of video and subtitle within a video. Given $\bm{\widetilde{V}}$ and $\bm{\widetilde{S}}$, the cross-modal representations are encoded as:
\begin{equation}
\begin{aligned}
    \bm{H'}_v & = \mathtt{Co\text{-}Transformer}_{vs}(\bm{\widetilde{V}},\;\bm{\widetilde{S}}) \\
    \bm{H'}_s & = \mathtt{Co\text{-}Transformer}_{sv}(\bm{\widetilde{S}},\;\bm{\widetilde{V}})
\end{aligned}
\label{eq:cross_transformer}
\end{equation}
where $\bm{H'}_v\in\mathbb{R}^{d\times n_v}$ and $\bm{H'}_s\in\mathbb{R}^{d\times n_v}$ are the learned cross-modal representations of video and subtitle, respectively.

Finally, we refine the encoded cross-modal representations of $\bm{H'}_v$ and $\bm{H'}_s$ with standard transformer blocks by learning the self-attentive contexts, respectively. The final output is calculated as:
\begin{equation}
    \bm{H}_v =\mathtt{Transformer}_{v}(\bm{H'}_v),\;\;
    \bm{H}_s =\mathtt{Transformer}_{s}(\bm{H'}_s)
\end{equation}
where $\bm{H}_v\in\mathbb{R}^{d\times n_v}$ and $\bm{H}_s\in\mathbb{R}^{d\times n_v}$ are the final output representations of video and subtitle, respectively.

If videos do not come with subtitles, then the feature encoding pipeline for subtitle will be removed. Accordingly, the co-attentional transformer becomes the standard transformer, and the final output is $\bm{H}_v$ only.

% =============================================
\subsection{Video Retrieval Module}\label{ssec:vr_module}
% =============================================
Through query encoding, a query is encoded to two $d$-dimensional vectors  $\bm{q}_m\in\mathbb{R}^{d}$,  $m\in\{v,s\}$, for matching with visual and subtitle features from a video. Recall that, with video encoding, each video is encoded to $\bm{H}_m=[\bm{h}_m^0,\bm{h}_m^2,\dots,\bm{h}_m^{n_v-1}]\in\mathbb{R}^{d\times n_v}$, \ie a sequence of $\bm{h}_m$'s each represents two $d$-dimensional vectors for visual and subtitle features extracted from a small fraction of a video.

We estimate the matching between the query and a video by cosine similarities computed on $\bm{q}_m$ and $\bm{H}_m$, \ie a simple late feature fusion. Specifically, we compute the cosine similarities between $\bm{q}_m$ and each element of $\bm{H}_m$ as:
\begin{equation}
    \bm{\varphi}_m=\mathtt{norm}(\bm{H}_m^{\top})\cdot\mathtt{norm}(\bm{q}_m)
\label{eq:vr_score}
\end{equation}
where $m\in\{v,s\}$, $\bm{\varphi}_m\in\mathbb{R}^{n_v}$, and $\mathtt{norm}$ denotes the $l2$ normalization operation. Then we select the maximum score in $\bm{\varphi}_m$ to represent the matching between query and video:
\begin{equation}
    \varphi_m=\max(\bm{\varphi}_m)=\max([\varphi_m^0,\varphi_m^1,\dots,\varphi_m^{n_v-1}])
\label{eq:max_vr_score}
\end{equation}
where $\varphi_m$ is a scalar. If videos come with subtitles, then $\varphi=\frac{1}{2}(\varphi_v + \varphi_s)$, otherwise $\varphi=\varphi_v$.

We adopt the hinge loss as training objective for video retrieval, similar to~\cite{yu2018mattnet,dong2019dual,gabeur2020multi,lei2020tvr}. We first sample two sets of negative pairs $\{(Q_i^{-},V)\}_{i=1}^{N}$ and $\{(Q,V_i^{-})\}_{i=1}^{N}$ for each positive pair $(Q,V)$, where $Q^-$ and $V^-$ denote the negative (\ie non-matching) query and video, respectively.\footnote{We simply use $V$ to represent a video with its subtitle if available.} Suppose the computed similarity scores of both sets of negative pairs are $\bm{\varphi'}$ and $\bm{\varphi''}$, the hinge loss is calculated as:
\begin{equation}
    \mathcal{L}^{VR}=\max(0,\Delta+\frac{1}{N}\sum\bm{\varphi'}-\varphi) + \max(0,\Delta+\frac{1}{N}\sum\bm{\varphi''}-\varphi)
\end{equation}
where $\Delta$ is the pre-defined margin value and we set $\Delta=0.1$.

% =============================================
\subsection{Moment Localization Module}\label{ssec:ml_module}
% =============================================
For efficiency purpose, moment localization is also computed based on the encoded query features  $\bm{q}_m$ and video features $\bm{H}_m$, through late feature fusion, following~\cite{lin2018bsn,lei2020tvr}. Specifically, $\bm{q}_m$ is further encoded with a feed-forward layer as $\bm{q'}_m=\bm{W}_m\cdot\bm{q}_m+\bm{b}_m\in\mathbb{R}^{d}$. Then we  compute  video-query similarity scores as:
\begin{equation}
    \bm{\mathcal{S}}_{mq}=\bm{H}_m^{\top}\cdot\bm{q'}_m\in\mathbb{R}^{n_v},\;\;\text{where}\;\; m\in\{v,s\}
\end{equation}
Again, if subtitle is available, $\bm{\mathcal{S}}=\frac{1}{2}(\bm{\mathcal{S}}_{vq}+\bm{\mathcal{S}}_{sq})$, otherwise $\bm{\mathcal{S}}=\bm{\mathcal{S}}_{vq}$. The start and end scores for target moment are generated by convolutional start-end boundary predictor~\cite{lei2020tvr}:
\begin{equation}
    \bm{\mathcal{S}}_{\text{start}}=\mathtt{Conv1D}_{\text{start}}(\bm{\mathcal{S}}),\;\;\bm{\mathcal{S}}_{\text{end}}=\mathtt{Conv1D}_{\text{end}}(\bm{\mathcal{S}})
\end{equation}
where $\bm{\mathcal{S}}_{\text{start/end}}\in\mathbb{R}^{n_v}$. Then, the probability distributions of start and end boundaries are computed by:
\begin{equation}
    \bm{P}_{\text{start}}=\mathtt{Softmax}(\bm{\mathcal{S}}_{\text{start}}),\;\;\bm{P}_{\text{end}}=\mathtt{Softmax}(\bm{\mathcal{S}}_{\text{end}})
\end{equation}
For a video-query pair, the predicted start and end boundaries of the target moment are derived by maximizing the joint probability:
\begin{equation}
\begin{aligned}
    (\hat{i}^s,\hat{i}^e) & = \arg\max_{a^s,a^e} \bm{P}_{\text{start}}(a^s)\times\bm{P}_{\text{end}}(a^e) \\
    P^{se} & = \bm{P}_{\text{start}}(\hat{i}^s)\times\bm{P}_{\text{end}}(\hat{i}^e)
\end{aligned}
\label{eq:st_ed_score}
\end{equation}
where $0\leq\hat{i}^s\leq \hat{i}^e\leq n_v - 1$, and $P^{se}$ is the score of best boundaries $(\hat{i}^s,\hat{i}^e)$. The training objective of moment localization is:
\begin{equation}
    \mathcal{L}^{ML} = \frac{1}{2}\times\Big(f_{\text{XE}}(\bm{P}_{\text{start}}, \bm{Y}_{\text{start}}) + f_{\text{XE}}(\bm{P}_{\text{end}}, \bm{Y}_{\text{end}})\Big)
\end{equation}
where $f_{\text{XE}}$ is the cross-entropy function, $\bm{Y}_{\text{start}}$ and $\bm{Y}_{\text{end}}$ are one-hot labels for start ($i^s$) and end ($i^e$) boundaries of the ground truth moment, respectively. 

We now have the  full picture of the base architecture ReLoNet with four modules: query encoder (Section~\ref{ssec:query_encoder}), video encoder (Section~\ref{ssec:video_encoder}), video retrieval and moment localization modules (Sections~\ref{ssec:vr_module} and~\ref{ssec:ml_module}). Next, we incorporate contrastive learning objectives into ReLoNet to develop ReLoCLNet.

% =============================================
\subsection{Video and Frame Contrastive Learning}\label{ssec:i2cl}
% =============================================
In ReLoNet, video retrieval and moment localization are fully based on the encoded query features $\bm{q}_m$ and video features $\bm{H}_m$. They are both computed by simple late future fusion. Quality of the final moment retrieval hence heavily relies on the effectiveness of the two separate encoders, query encoder and video encoder.

In ReLoCLNet, we aim to guide the two encoders to simulate cross-modal interaction learning in the training phase. To this end, we introduce two contrastive learning objectives, VideoCL and FrameCL. VideoCL guides the two encoders to better distinguish matching video-query pairs from non-matching pairs. FrameCL guides the two encoders to better distinguish the matching moment to the query from the non-matching moments.

% =============================================
\subsubsection{Video Contrastive Learning (VideoCL)}\label{sssec:inter_cl}
% =============================================
VideoCL guides the encoders to learn a joint feature space where the semantically related videos and queries are close to each other, and far away otherwise. In other words, VideoCL aims to reduce the distance of matching video-query pairs, and to increase the distance of non-matching pairs, in the joint feature space. 

We encode the latent representation of video $\bm{H'}_m\in\mathbb{R}^{d\times n_v}$ from Eq.~\ref{eq:cross_transformer} (illustrated as  $\bm{H'}_v$ and  $\bm{H'}_s$ in Figure~\ref{fig:architecture}) into its modularized video representation $\bm{c}_m$. Similar to modular component in query encoder, we adopt additive attention mechanism to compute $\bm{c}_m$:
\begin{equation}
    \bm{\alpha}^m = \mathtt{Softmax}\big(\bm{W}_{m,\alpha}\cdot\bm{H'}_m\big) \in \mathbb{R}^{n_v},\;\;
    \bm{c}_{m} = \sum_{i=0}^{n_v-1}\alpha_i^m\times\bm{h'}_{m,i}
\end{equation}
where $\bm{c}_m\in\mathbb{R}^{d}$, $\bm{W}_{m,\alpha}\in\mathbb{R}^{1\times d}$ and $m\in\{v,s\}$.

Given a set of positive (\ie matching) video-query pairs $\mathcal{P}=\{(\bm{c}_m,\bm{q}_m)\}$ and the sampled set of negative (\ie non-matching) video-query pairs $\mathcal{N}=\{(\bm{c'}_m,\bm{q'}_m)\}$, we adopt the noise-contrastive estimation (NCE)~\cite{michael2010noise,jozefowicz2016exploring,sun2019cbt,miech2020end} to compute the VideoCL score:
\begin{equation}
    \mathcal{I}_m^e=\log\Bigg(
    \frac{
    \sum\limits_{(\bm{c}_m,\bm{q}_m)\in\mathcal{P}}e^{f(\bm{c}_m)^{\top}\cdot g(\bm{q}_m)}}
    {
    \sum\limits_{(\bm{c}_m,\bm{q}_m)\in\mathcal{P}}e^{f(\bm{c}_m)^{\top}\cdot g(\bm{q}_m)} 
    + 
    \sum\limits_{(\bm{c'}_m,\bm{q'}_m)\sim\mathcal{N}}e^{f(\bm{c'}_m)^{\top}\cdot g(\bm{q'}_m)}
    }\Bigg)
\label{eq:nce_score}
\end{equation}
where the exponential term, $e^{f(\bm{c})^{\top}\cdot g(\bm{q})}$, computes the mutual information (MI) between $\bm{c}$ and $\bm{q}$. $f(\cdot)$ and $g(\cdot)$ denote the parametrized mappings, which project video and query representations into the same embedding space. Again,  $\mathcal{I}^e=\frac{1}{2}(\mathcal{I}_v^e + \mathcal{I}_s^e)$ if subtitle is available, otherwise $\mathcal{I}^e=\mathcal{I}_v^e$. 

The objective of NCE is to optimize $\max_{f,g}(\mathcal{I}^e)$, which is equivalent to maximizing the ratio of the summed MI's of all samples in $\mathcal{P}$ and the summed MI's of all samples in $\mathcal{N}$~\cite{miech2020end}. The loss of VideoCL is defined as:
\begin{equation}
    \mathcal{L}^{VideoCL}=-\mathcal{I}^e
\end{equation}

\begin{figure}[t]
    \centering
    \includegraphics[trim={0.35cm 0cm 1.05cm 0cm},clip,width=0.7\linewidth]{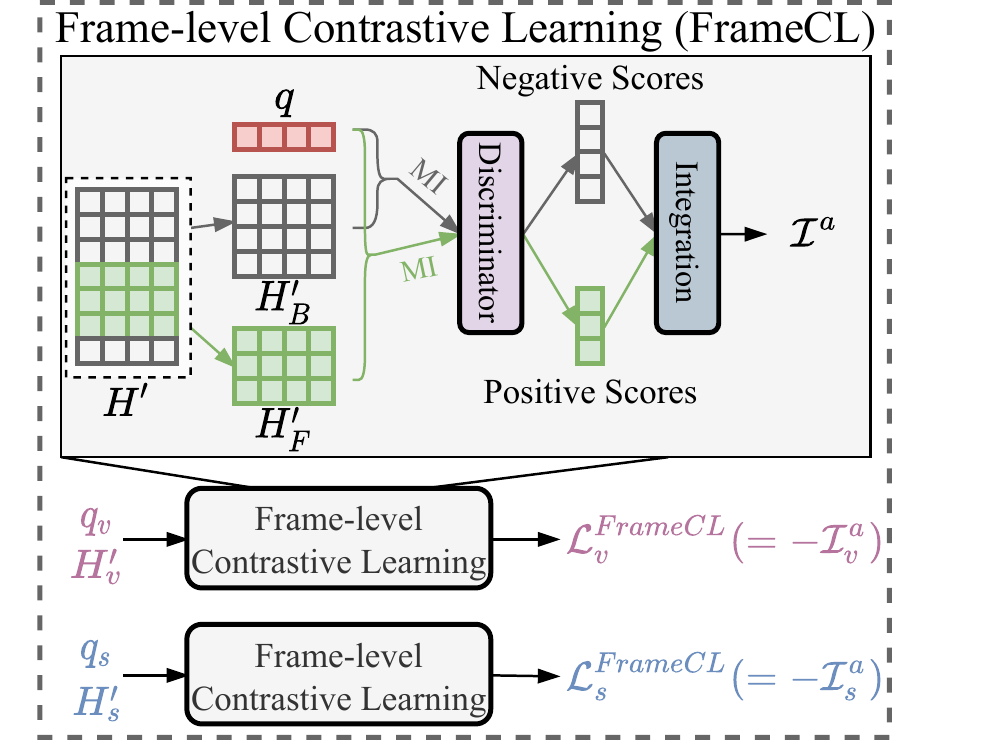}
    \caption{Structure of the FrameCL module.}
	\label{fig:intra_cl}
\end{figure}

% =============================================
\subsubsection{Frame Contrastive Learning (FrameCL)}\label{sssec:intra_cl}
% =============================================
FrameCL focuses on moment localization within a given pair of video-query, where the video retrieval module predicts the video contains a matching moment to the query. We regard the video features that reside within boundaries of the target moment as foreground or positive samples, and the rest as background or negative samples. Then we compute the contrastive loss by measuring MI between the query and the positive/negative video features. For this purpose, we utilize a discriminative approach based on mutual information maximization~\cite{hjelm2018learning,velickovic2018deep}.

The structure of FrameCL module is shown in Figure~\ref{fig:intra_cl}. The inputs $\bm{H'}_v$, $\bm{H'}_s$, $\bm{q}_v$, and $\bm{q}_s$ are outputs illustrated in Figure~\ref{fig:architecture}. Given the latent representation of video $\bm{H'}_m\in\mathbb{R}^{d\times n_v}$, we first split it into two parts by boundaries of target moment. The positive/foreground video features are $\bm{H'}_{m,F}=\{\bm{h'}_{m,i}|i=i^s,\dots,i^e\}\in\mathbb{R}^{d\times n_t}$, which are features within the target moment.\footnote{$n_t=i^e-i^s+1$, and it denotes the length of target moment.} The negative/background features  $\bm{H'}_{m,B}=\{\bm{h'}_{m,i}|i=0,\dots,i^s-1,i^e+1,\dots,n_v-1\}\in\mathbb{R}^{d\times(n_v-n_t)}$, are not in the target moment. 

With query representation $\bm{q}_m$, foreground representation $\bm{H'}_{m,F}$, and background representation $\bm{H'}_{m,B}$, our goals are to maximize the MI between the query and the foreground, as well as to minimize the MI between the query and the background. Since MI estimation is in general intractable for continuous and random variables, we choose to maximize the value over lower bound estimators of MI, through Jensen-Shannon MI estimator~\cite{hjelm2018learning} as:
\begin{equation}
\begin{aligned}
    \mathcal{I}_m^a = \mathbb{E}_{\bm{H'}_{m,F}}\Big[-\mathtt{sp}\big(-\mathcal{C}_{\theta}(\bm{q},\bm{H'}_{m,F})\big)\Big] - \mathbb{E}_{\bm{H'}_{m,B}}\Big[\mathtt{sp}\big(\mathcal{C}_{\theta}(\bm{q},\bm{H'}_{m,B})\big)\Big]
\end{aligned}
\end{equation}
where $\mathtt{sp}(x)=\log(1+ex)$ is the Softplus activation. $\mathcal{C}_{\theta}:d\times d\to\mathbb{R}$ refers to a discriminator. Similarly, $\mathcal{I}^a=\frac{1}{2}(\mathcal{I}_v^a+\mathcal{I}_s^a)$ if subtitle is available, otherwise $\mathcal{I}^a=\mathcal{I}_v^a$. The contrastive loss of FrameCL is:
\begin{equation}
    \mathcal{L}^{FrameCL}=-\mathcal{I}^a
\end{equation}

Note that, both VideoCL and FrameCL are training objectives, and their losses are used to update video and query encoders. Although the two objectives are designed for video retrieval and moment localization respectively, they mutually affect each other, because both video and query encoders are adjusted based on the loss from both VideoCL and FrameCL, together with other losses. 

% =============================================
\subsection{Training and Inference}\label{ssec:train_infer}
% =============================================
The overall training loss for ReLoCLNet is:
\begin{equation}
    \mathcal{L}=\lambda_1\times\mathcal{L}^{VR} + \lambda_2\times\mathcal{L}^{ML} + \lambda_3\times\mathcal{L}^{VideoCL} + \lambda_4\times\mathcal{L}^{FrameCL}
\end{equation}
$\lambda_i$'s are hyperparameters to balance the contribution of each loss. We set $\lambda_1=1.0$ and $\lambda_{2,3,4}=0.01$ to keep all losses at the same order of magnitude \ie equal contributions from the four components. Note that each video contains a large number of candidate moments.

During inference for VCMR, given a text query and a video corpus with $M$ videos, we first use Eq.~\ref{eq:vr_score} and~\ref{eq:max_vr_score} to compute the similarity between the query and each of the $M$ videos, leading to $\bm{\varphi}=[\varphi_1,\varphi_2,\dots,\varphi_M]$. The top-$K$ most relevant  videos are retrieved based on $\bm{\varphi}$ ($K=100$ in our implementation). For each retrieved video, we compute the scores of a few candidate predicted moments by Eq.~\ref{eq:st_ed_score}. Let $P^{se}$ be the score of one predicted moment in the video. The final VCMR score is computed by:
\begin{equation}
    \delta = P^{se}\times e^{\gamma\cdot\varphi}
\end{equation}
The exponential term and the hyperparameter $\gamma$ are used to balance the importance of video retrieval and moment localization scores. 

\begin{table}
  \caption{The hyper-parameters for TVR and ANetCaps}
  \label{tab:hyperparam}
  \centering
  \begin{tabular}{| l | c | c |}
    \specialrule{0.08em}{0.0em}{0.0em}
    Hyperparameter Name & TVR & ANetCaps\\
    \hline
    $n_v$ (max video sequence) & \multicolumn{2}{c|}{128} \\
    \hline
    $n_q$ (max query sequence) & 30 & 64 \\
    \hline
    $d_v$ (visual feature dim) & 3072~\scriptsize{2048(ResNet)+1024(I3D)}  & 1024~\scriptsize{(I3D)} \\
    \hline
    $d_w$ (word feature dim) & 768~\scriptsize{(RoBERTa)} & 300~\scriptsize{(GloVe)} \\
    \hline
    $d$ (hidden size) & \multicolumn{2}{c|}{384} \\
    \hline
    $\gamma$ & 30 & 20 \\
    \hline
    \multicolumn{3}{|p{8cm}|}{
     \# negative samples in VR: 10 \hfill Optimizer: AdamW~\cite{devlin2019bert} \newline 
     Dropout rate: 0.1\hfill  Weight decay rate: 0.01  \hfill Batch size: 128 \newline
     Learning rate (lr): 0.0001 \hfill lr warmup proportion: 0.01 \newline
    Early stop tolerance: 10\hfill    \# total training epochs: 100
    } \\
    \specialrule{0.08em}{0.0em}{0.0em} 
\end{tabular}
\end{table}

% =============================================
\section{Experiments}\label{sec:exp}
% =============================================

% =============================================
\subsection{Datasets and Evaluation Metrics}\label{ssec:dataset}
% =============================================
We conduct experiments on two benchmark datasets: ActivityNet Captions~\cite{krishna2017dense} and TVR~\cite{lei2020tvr}. ActivityNet Captions (\textbf{ANetCaps}) contains around $20$K videos taken from the ActivityNet~\cite{heilbron2015activitynet} dataset. The average video duration is about $120$ seconds, the average query length is around $14.78$ words, the average moment duration is about $36.18$ seconds, and each video contains $3.68$ annotations on average. This dataset is originally designed for SVMR task, then adapted to VCMR by~\citet{escorcia2019temporal}. We follow the setup in~\cite{escorcia2019temporal,zhang2020hierarchical} with $10,009$ and $4,917$ videos (\ie $37,421$ and $17,505$ annotations) for train and test, respectively. \textbf{TVR} is collected by \citet{lei2020tvr}, which contains $21.8$K videos and $109$K queries in total. The average video duration is $76.2$ seconds, the query contains $13.4$ words on average, the average moment duration is $9.1$ seconds, and each video contains $5$ annotations on average. We follow~\citet{zhang2020hierarchical} with $17,435$ and $2,179$ videos for train and test, respectively. Same as~\citet{lei2020tvr} and~\citet{zhang2020hierarchical}, we utilize both video and subtitle features in the TVR dataset for train and test.

We evaluate the models for the VCMR task as well as its two subtasks: video retrieval (VR) and SVMR. For VR, we use ``\textbf{Recall$@k$}'' ($k\in\{1,5,10,100\}$) as the evaluation metric following~\cite{lei2020tvr,zhang2020hierarchical}. Note that we do not use ``Precision@$k$'' because each query only corresponds to one ground truth video, in both datasets. For SVMR and VCMR, we use ``\textbf{$\text{Recall@}k,\text{IoU=}\mu$}'' as the evaluation metric, which denotes the percentage of test samples that have at least one predicted moment whose \textit{intersection over union} (IoU) with the ground-truth moment is larger than $\mu$ in the top-$k$ predictions. We set $k\in\{1,10,100\}$ and $\mu\in\{0.5,0.7\}$. A prediction is correct if (i) the predicted video matches the ground truth video, and (ii) the predicted moment has high overlap with the ground truth moment, where temporal IoU is used to measure the overlap~\cite{lei2020tvr}.

% =============================================
\subsection{Implementation Details}\label{ssec:impl}
% =============================================
For ANetCaps, we use I3D~\cite{Carreira2017QuoVA} pre-trained on Kinetics dataset~\cite{kay2017kinetics} as the visual feature extractor following~\citet{zhang2020hierarchical}, and adopt GloVe embeddings~\cite{pennington2014glove} as the textual feature extractor for query words. For TVR, we directly use the visual and textual features provided by~\citet{lei2020tvr}. The visual feature is the concatenation of appearance feature extracted by ResNet152~\cite{he2016resnet} pre-trained on ImageNet~\cite{deng2009imagenet} and temporal feature extracted by I3D. The textual feature of query and subtitle is extracted by 12-layer pre-trained RoBERTa~\cite{liu2019roberta}. The negative sets of video retrieval and VideoCL modules are sampled within each mini-batch during training. The hyperparameters are summarized in Table~\ref{tab:hyperparam}. Other hyperparameters are given when describing the corresponding model components. Our model is implemented in PyTorch $\mathtt{1.7.0}$ with CUDA $\mathtt{11.1}$ and cudnn $\mathtt{8.0.5}$. All experiments are conducted on a workstation with dual NVIDIA GeForce RTX 3090 GPUs.

\begin{table}[t]
	\centering
	\caption{\small Results of VCMR on TVR and ANetCaps datasets}
	\label{tab:vcmr_sota}
	\setlength{\tabcolsep}{1.8pt}
	\begin{tabular}{c | l | c c c | c c c }
		\specialrule{.1em}{.05em}{.05em}
		\multirow{2}{*}{Dataset} & \multirow{2}{*}{Method} & \multicolumn{3}{c |}{$\text{Recall@}k, \text{IoU}=0.5$} & \multicolumn{3}{c}{$\text{Recall@}k, \text{IoU}=0.7$} \\
        & & R1 & R10 & R100 & R1 & R10 & R100 \\
        \hline
        \multirow{6}{*}{\tabincell{l}{\rotatebox{90}{TVR}}}
        & XML~\cite{lei2020tvr}      & -    & -     & -     & 2.62 & 9.05  & 22.47 \\
        & HERO~\cite{li2020hero}     & -    & -     & -     & 2.98 & 10.65 & 18.25 \\
        & FLAT~\cite{zhang2020hierarchical}     & \textit{8.45} & 21.14 & 30.75 & \textit{4.61} & 11.29 & 16.24 \\
        & HAMMER~\cite{zhang2020hierarchical}   & \textbf{9.19} & \textit{21.28} & 31.25 & \textbf{5.13} & \textit{11.38} & 16.71 \\
        \cline{2-8}
        & ReLoNet  & 5.46 & 16.65 & \textit{35.08} & 2.71 & 9.37  & \textit{22.87} \\
        & ReLoCLNet  & 8.03 & \textbf{21.37} & \textbf{44.10} & 4.15 & \textbf{14.06} & \textbf{32.42} \\
        \specialrule{.1em}{.05em}{.05em}
        \multirow{6}{*}{\tabincell{l}{\rotatebox{90}{ANetCaps}}}
        & MCN~\cite{hendricks2017localizing}      & 0.02 & 0.18  & 1.26  & 0.01 & 0.09 & 0.70 \\
        & CAL~\cite{escorcia2019temporal}      & 0.21 & 1.32  & 6.82  & 0.12 & 0.89 & 4.79 \\
        & FLAT~\cite{zhang2020hierarchical}     & 2.57 & \textit{13.07} & \textit{30.66} & 1.51 & \textit{7.69} & 17.67 \\
        & HAMMER~\cite{zhang2020hierarchical}   & \textit{2.94} & \textbf{14.49} & \textbf{32.49} & \textit{1.74} & \textbf{8.75} & \textbf{19.08} \\
        \cline{2-8}
        & ReLoNet  & 2.16 & 9.96 & 24.54 & 1.26 & 5.64 & 17.43 \\
        & ReLoCLNet  & \textbf{3.09} & 11.28 & 25.95 & \textbf{1.82} & 6.91 & \textit{18.33} \\
        \specialrule{.1em}{.05em}{.05em}
	\end{tabular}
\end{table}

\begin{table}[t]
	\centering
	\caption{\small Retrieval efficiency on the TVR dataset}
	\label{tab:retrieval_time}
	\begin{tabular}{l | r r}
		\specialrule{.1em}{.05em}{.05em}
		\multirow{2}{*}{Method} & \multicolumn{2}{c}{Retrieval Efficiency}  \\
        & Total Time & Average Per Query \\
        \hline
        XML~\cite{lei2020tvr}      & $39.34$ seconds & $3.61$ milliseconds \\
        HAMMER~\cite{zhang2020hierarchical}   & $2,378.67$ seconds & $218.33$ milliseconds \\
        \hline
        ReLoNet  & \multirow{2}{*}{$42.07$ seconds} & \multirow{2}{*}{$3.86$ milliseconds} \\
        ReLoCLNet &  &  \\
        \specialrule{.1em}{.05em}{.05em}
	\end{tabular}
\end{table}

\begin{table}[t]
	\centering
	\caption{\small Results of VR subtask on TVR and ANetCaps datasets}
	\label{tab:vr_sota}
	\setlength{\tabcolsep}{4.5pt}
	\begin{tabular}{c | l | c c c c}
		\specialrule{.1em}{.05em}{.05em}
		\multirow{2}{*}{Dataset} & \multirow{2}{*}{Method} & \multicolumn{4}{c}{$\text{Recall@}k$} \\
        & & $k=1$ & $k=5$ & $k=10$ & $k=100$ \\
        \hline
        \multirow{6}{*}{TVR}
        & MCN~\cite{hendricks2017localizing}      & 0.05  & 0.38  & 0.66  & 3.59  \\
        & CAL~\cite{escorcia2019temporal}      & 0.28  & 1.02  & 1.68  & 8.55  \\
        & MEE~\cite{miech2018learning}      & 7.56  & 20.78 & 29.88 & 73.07 \\
        & XML~\cite{lei2020tvr}      & 16.54 & 38.11 & 50.41 & 88.22 \\
        \cline{2-6}
        & ReLoNet  & \textit{16.96} & \textit{39.28} & \textit{51.34} & \textit{88.46} \\
        & ReLoCLNet  & \textbf{22.13} & \textbf{45.85} & \textbf{57.25} & \textbf{90.21} \\
        \specialrule{.1em}{.05em}{.05em}
        \multirow{4}{*}{ANetCaps}
        & FLAT~\cite{zhang2020hierarchical}     & 5.37 & - & 29.14 & 71.64 \\
        & HAMMER~\cite{zhang2020hierarchical}   & 5.89 & - & 30.98 & 73.38 \\
        \cline{2-6}
        & ReLoNet  & \textit{7.02} & \textit{24.42} & \textit{35.24} & \textit{78.08} \\
        & ReLoCLNet  & \textbf{9.64} & \textbf{28.02} & \textbf{40.26} & \textbf{79.13} \\
        \specialrule{.1em}{.05em}{.05em}
	\end{tabular}
\end{table}

\begin{table}[t]
	\centering
	\caption{\small Results of SVMR subtask on TVR and ANetCaps datasets}
	\label{tab:svmr_sota}
	\begin{tabular}{c | l | c c c}
		\specialrule{.1em}{.05em}{.05em}
		\multirow{2}{*}{Dataset} & \multirow{2}{*}{Method} & \multicolumn{3}{c}{$\text{Recall@}1, \text{IoU}=\mu$} \\
        & & $\mu=0.3$ & $\mu=0.5$ & $\mu=0.7$ \\
        \hline
        \multirow{6}{*}{TVR}
        & MCN~\cite{hendricks2017localizing}      & -     & 13.08 & 5.06  \\
        & CAL~\cite{escorcia2019temporal}      & -     & 12.07 & 4.68  \\
        & ExCL~\cite{ghosh2019excl}     & -     & \textit{31.34} & \textit{14.19} \\
        & XML~\cite{lei2020tvr}      & -     & 30.75 & 13.41 \\
        \cline{2-5}
        & ReLoNet  & \textit{48.14} & 29.49 & 13.13 \\
        & ReLoCLNet  & \textbf{49.87} & \textbf{31.88} & \textbf{15.04} \\
        \specialrule{.1em}{.05em}{.05em}
        \multirow{4}{*}{ANetCaps}
        & FLAT~\cite{zhang2020hierarchical}     & \textit{57.58} & \textit{39.60} & \textit{22.59} \\
        & HAMMER~\cite{zhang2020hierarchical}   & \textbf{59.18} & \textbf{41.45} & \textbf{24.27} \\
        \cline{2-5}
        & ReLoNet  & 39.27 & 23.67 & 14.55 \\
        & ReLoCLNet  & 42.65 & 28.54 & 17.76 \\
        \specialrule{.1em}{.05em}{.05em}
	\end{tabular}
\end{table}

\begin{table*}[t]
	\centering
	\caption{\small The effects of different objectives on TVR dataset (VR=Video Retrieval, ML=Moment Localization, VideoCL=Video Contrastive Learning, and FrameCL=Frame Contrastive Learning)}
	\label{tab:component_ablation}
	\setlength{\tabcolsep}{3.0 pt}
	\begin{tabular}{c c c c | c c c | c c c | c c c | c c c | c c c }
		\specialrule{.1em}{.05em}{.05em}
		\multicolumn{4}{c |}{\multirow{2}{*}{Objective}} 
		& \multicolumn{6}{c |}{VCMR} 
		& \multicolumn{3}{c |}{VR} 
		& \multicolumn{6}{c}{SVMR}  \\
		\cline{5-19}
		& & & 
		& \multicolumn{3}{c |}{$\text{Recall@}k, \text{IoU=}0.5$} 
		& \multicolumn{3}{c |}{$\text{Recall@}k, \text{IoU=}0.7$} 
		& \multicolumn{3}{c |}{$\text{Recall@}k$} 
		& \multicolumn{3}{c |}{$\text{Recall@}k, \text{IoU=}0.5$} 
		& \multicolumn{3}{c}{$\text{Recall@}k, \text{IoU=}0.7$}  \\
		\cline{1-19}
		VR & ML & VideoCL & FrameCL & 1 & 10 & 100 & 1 & 10 & 100 & 1 & 10 & 100 & 1 & 10 & 100 & 1 & 10 & 100 \\
        \hline
        \ding{52} & \ding{56} & \ding{56} & \ding{56} 
        & - & - & - & - & - & - & 16.23 & 49.33 & 87.38 & - & - & - & - & - & - \\
        \ding{56} & \ding{52} & \ding{56} & \ding{56} 
        & - & - & - & - & - & - & - & - & - & 30.21 & 59.81 & 83.43 & 13.91 & 41.55 & 68.51 \\
        \ding{52} & \ding{52} & \ding{56} & \ding{56}
        & 5.46 & 16.65 & 35.08 & 2.71 & 9.37 & 22.87 & 16.96 & 51.34 & 88.46 & 29.49 & 54.06 & 75.89 & 13.13 & 35.46 & 58.84 \\
        \hline
        \ding{52} & \ding{52} & \ding{52} & \ding{56}
        & 6.63 & 18.16 & 39.69 & 3.24 & 11.78 & 27.69 & \textit{20.69} & \textit{55.70} & \textit{89.71} & 29.52 & 57.32 & 78.65 & 13.76 & 38.26 & 64.27 \\
        \ding{52} & \ding{52} & \ding{56} & \ding{52} 
        & \textit{7.21} & \textit{20.04} & \textit{42.45} & \textit{3.75} & \textit{12.77} & \textit{30.32} & 19.81 & 54.38 & 88.96 & \textit{31.75} & \textit{62.20} & \textit{85.99} & \textit{14.73} & \textit{44.60} & \textit{71.44} \\
        \ding{52} & \ding{52} & \ding{52} & \ding{52}
        & \textbf{8.03} & \textbf{21.37} & \textbf{44.10} & \textbf{4.15} & \textbf{14.06} & \textbf{32.42} & \textbf{22.13} & \textbf{57.25} & \textbf{90.21} & \textbf{31.88} & \textbf{63.89} & \textbf{86.67} & \textbf{15.04} & \textbf{45.24} & \textbf{72.12} \\
        \specialrule{.1em}{.05em}{.05em}
	\end{tabular}
\end{table*}

% =============================================
\subsection{Performance Comparison}\label{ssec:performance}
% =============================================
We compare our models with MCN~\cite{hendricks2017localizing}, CAL~\cite{escorcia2019temporal}, XML~\cite{lei2020tvr}, HERO~\cite{li2020hero}, FLAT~\cite{zhang2020hierarchical} and HAMMER~\cite{zhang2020hierarchical}. Among them, MCN, CAL, XML, and HERO follow unimodal encoding approaches, while FLAT and HAMMER belong to cross-modal interaction learning approaches (see Figure~\ref{fig:interaction_example}). FLAT is a variant of HAMMER without using hierarchical structure. In all tables, results of the compared models are reported in their corresponding papers.\footnote{Two sets of results are reported for HERO in~\cite{li2020hero}, with and without large-scale pre-training. We choose the version without pre-training as all other models compared here do not use pre-training.} The best results are in \textbf{boldface} and the second bests are in \textit{italic}.

The results of VCMR on TVR and ANetCaps datasets are reported in Table~\ref{tab:vcmr_sota}. On TVR dataset, ReLoNet is comparable to XML with slightly better performance. ReLoCLNet outperforms all baselines over Recall@$10$ and Recall@$100$ metrics. Observe that the performance of ReLoCLNet is lower than FLAT and HAMMER over Recall@$1$. Since both FLAT and HAMMER adopt fine-grained cross-modal interaction learning, they are more adequate to align video and query for precise moment retrieval. Compared with ReLoNet, ReLoCLNet achieves significant improvements over all evaluation metrics, which demonstrate the effectiveness of the proposed contrastive learning objectives. 

On ANetCaps dataset, ReLoNet surpasses the ranking-based methods, MCN and CAL, by large margins over all evaluation metrics. Similarly, ReLoCLNet is  superior to ReLoNet thanks to the contrastive learning components. Compared with FLAT and HAMMER, ReLoCLNet outperforms both over Recall@$1$ but is poorer over Recall@$10$ and Recall@$100$. This observation is contrary to that on TVR dataset. Recall that FLAT and HAMMER adopt cross-modal interactions learning between video and query, and we have separate encoders for video and query. In addition, FLAT and HAMMER utilize pre-trained RoBERTa to extract textual features for query, while we simply adopt GloVe embeddings. All these contribute the differences between our results and that of  FLAT and HAMMER. Overall, we consider ReLoCLNet achieves comparable effectiveness with FLAT and HAMMER.

\begin{figure}[t]
    \centering
    \includegraphics[trim={0cm 7.45cm 0cm 7.35cm},clip,width=0.9\linewidth]{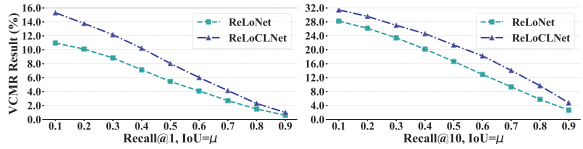}
    \caption{Recall@$1$ and Recall@$10$  of VCMR on TVR dataset over different IoU thresholds.}
	\label{fig:vcmr_iou}
\end{figure}

\begin{figure}[t]
    \centering
    \includegraphics[trim={0cm 7.45cm 0cm 7.35cm},clip,width=0.9\linewidth]{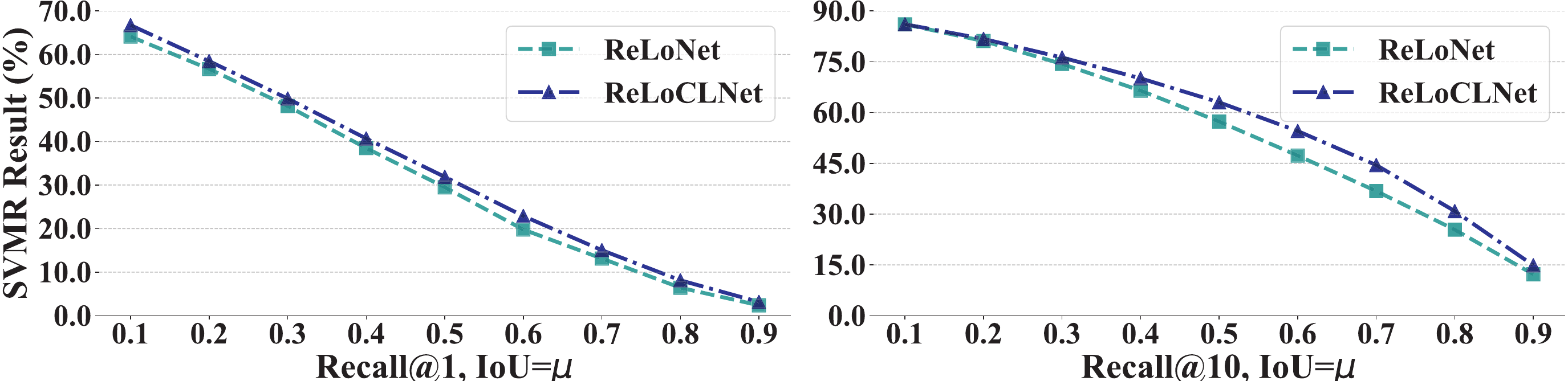}
    \caption{Recall@$1$ and Recall@$10$ of SVMR on TVR dataset over different IoU thresholds.}
	\label{fig:svmr_iou}
\end{figure}

\begin{figure}[t]
    \centering
    \includegraphics[trim={0cm 5cm 0cm 5cm},clip,width=0.9\linewidth]{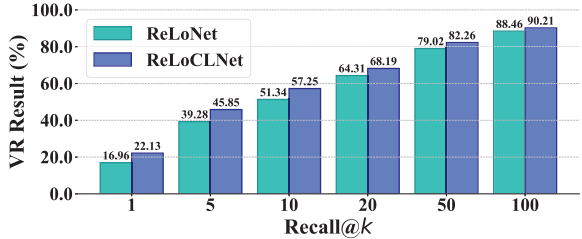}
    \caption{Recall@$K$ of VR on TVR dataset over different $K$.}
	\label{fig:vr_threshold}
\end{figure}

\begin{figure*}[t]
    \centering
    \includegraphics[trim={0cm 0cm 0cm 0cm},clip,width=0.85\textwidth]{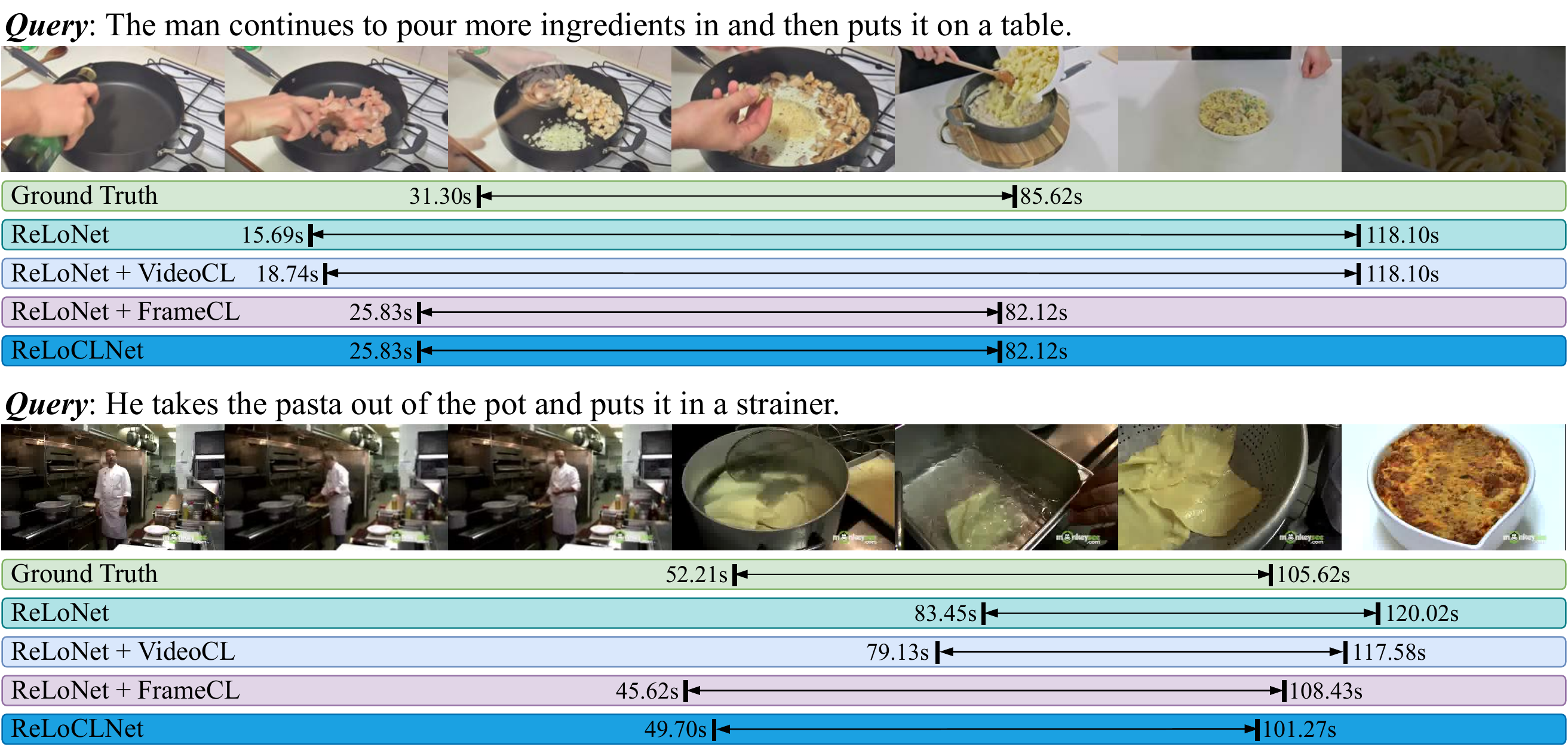}
    \caption{Visualization of moment localization predictions by ReLoNet, ReLoCLNet, and ReLoNet with VideoCL or FrameCL, for two queries on ANetCaps dataset.}
	\label{fig:visualization}
\end{figure*}

% =============================================
\subsection{Retrieval Efficiency and Ablation Study}\label{ssec:analysis}
% =============================================
In this section, we compare retrieval efficiency and perform in-depth ablation studies. We study the performance of our models on VR and SVMR subtasks, and the effects of different components.

% =============================================
\subsubsection{Retrieval Efficiency}\label{sssec:efficiency}
% =============================================
We consider VCMR in the validation set of TVR dataset containing $2,179$ videos with $10,895$ queries. The retrieval efficiency is summarized in Table~\ref{tab:retrieval_time}. The time spent on data pre-processing and feature extraction by pre-trained extractor are not counted since the same process applies to all methods. We used the XML code released by the authors, and re-implemented HAMMER according to their paper as its code is not released. Observe that the retrieval efficiency of our models are comparable to XML, and our models are far more efficient than HAMMER. Although HAMMER performs better on more strict metrics (\eg Recall@$1$, IoU=$0.7$), our models are around $56.71$ times faster than HAMMER in retrieval. Note that, ReLoCLNet and ReLoNet have the same retrieval efficiency, because neither VideoCL nor FrameCL introduces additional parameters; and all additional computations of ReLoCLNet happen in training stage.

% =============================================
\subsubsection{Video Retrieval Subtask}\label{sssec:vr}
% =============================================
Table~\ref{tab:vr_sota} reports the results on TVR and ANetCaps datasets. Observe that ReLoNet performs slightly better than XML on TVR, and significantly better than HAMMER on ANetCaps. ReLoCLNet outperforms all baselines by large margins on both datasets. In particular, ReLoCLNet achieves $5.59\%$ improvement in Recall@$1$ comparing with XML on TVR dataset. On ANetCaps dataset, ReLoCLNet obtains $9.64\%$ absolute score in Recall@$1$, compared with $5.89\%$ of HAMMER. 

% =============================================
\subsubsection{Single Video Moment Retrieval Subtask}\label{sssec:svmr}
% =============================================
The results of SVMR  on both datasets are reported in Table~\ref{tab:svmr_sota}. On TVR, ReLoCLNet achieves best performance, and obtains significant improvements against ReLoNet. Compared with ExCL, ReLoCLNet only outperforms by a small margin.  ExCL is specially designed for SVMR, with fine-grained cross-modal interactions learning. On ANetCaps, ReLoCLNet is superior to ReLoNet by large margins, which again shows the effectiveness of contrastive learning. However, ReLoCLNet performs worse than FLAT and HAMMER. Because both FLAT and HAMMER inherit their architectures designed for SVMR, which contain sophisticated and computational expensive cross-modal interactions for high-quality moment retrieval. In contrast, ReLoCLNet only relies on simple late fusion of separately encoded query and video features.

% =============================================
\subsubsection{Analysis on the Learning Objectives}\label{sssec:objective}
% =============================================
Table~\ref{tab:component_ablation} reports the contributions of different training objectives on TVR dataset. Note ReLoNet equals to VR+ML objectives, and ReLoCLNet is with all the four objectives. We first analyze the video retrieval (VR) and moment localization (ML) objectives. ReLoNet jointly trains VR and ML objectives for the VCMR task. Comparing VR with ReLoNet, the performance of ReLoNet on video retrieval is slightly better than that of VR, which means the ML objective also contributes to refine video retrieval learning process. In contrast, compared to ML only, ReLoNet underperforms ML on moment localization with marginal performance degradation, which implies that VR objective has negligible negative impact on moment localization.

Now we analyze the effects of VideoCL and FrameCL objectives. Observe that VideoCL contributes to performance improvements on both VCMR and VR, while it achieves marginal improvements on SVMR. Recall that VideoCL adopts noise-contrastive estimation to enlarge the similarities of matched video-query pairs, and reduce similarities between unpaired videos and queries; this is in line with video retrieval objective. Thus, it is beneficial to video retrieval learning. ReLoNet with FrameCL outperforms ReLoNet on all the three tasks. FrameCL aims to distinguish the matching moment from non-matching moment within a video. In this case, FrameCL guides the model to search for boundaries of target moment for precise moment localization. In fact, the matching between query and video is largely based on the matching moment in the video. In this sense, by highlighting matching moment, FrameCL does contribute to video retrieval task as well. Combining VideoCL and FrameCL, ReLoCLNet further boosts the performances on all three tasks by incorporating the advantages of both VideoCL and FrameCL.

% =============================================
\subsection{Qualitative Analysis}\label{ssec:qual}
% =============================================
Figure~\ref{fig:vcmr_iou} plots Recall@$1$ and Recall@$10$ of VCMR performances on TVR dataset over different IoU thresholds. We evaluate $9$ different IoU($\mu$) values, from $0.1$ to $0.9$. ReLoCLNet consistently outperforms ReLoNet, and relative performance improvements of ReLoCLNet are larger under more strict metrics. For instance, compared with ReLoNet, ReLoCLNet achieves $47.07\%$ relative gains ($8.03$ vs $5.46$) in Recall@$1$, IoU=$0.5$ versus $28.35\%$ relative gains ($21.37$ vs $16.65$) in Recall@$10$, IoU=$0.5$. 

Figure~\ref{fig:svmr_iou} plots Recall@$1$ and Recall@$10$  of SVMR over different IoU thresholds, and similar observations hold on this task. Figure~\ref{fig:vr_threshold} plots the video retrieval (VR) results of ReLoNet and ReLoCLNet over different recall thresholds on TVR dataset. Similarly, ReLoCLNet surpasses ReLoNet over all thresholds, and the relative performance improvement ratio is larger under more strict metrics.

Finally, we show two retrieval examples in Figure~\ref{fig:visualization} from ANetCaps dataset. The figure shows the predicted moments by ReLoCLNet and ReLoNet+FrameCL are closer to ground truth than that by ReLoNet and ReLoNet+VideoCL, which demonstrates the effectiveness of FrameCL module. Note FrameCL is designed to maximize the mutual information between query and frames within the target moment, and to minimize the MI between the query and frames outside target moment. With FrameCL, the model is guided to search for the boundaries within the region of target moment.

% =============================================
\section{Conclusion}\label{sec:conclusion}
% =============================================
In this paper, we propose a Retrieval and Localization Network with Contrastive Learning (ReLoCLNet) for video corpus moment retrieval (VCMR) task. Specifically, we introduce two contrastive learning objectives (VideoCL and FrameCL) on top of a unimodal encoding approach, ReLoNet, to address the contradiction between retrieval efficiency and retrieval quality. The VideoCL objective guides the video and query encoders to shorten the distance of matching videos and queries while enlarge the non-matching pairs. The FrameCL objective works at frame-level to simulate the fine-grained cross-modal interactions between visual and textual features within a video. Through extensive experimental studies, we show that ReLoCLNet addresses VCMR with high efficiency, and its retrieval accuracy is comparable with state-of-the-art methods which are much costly in terms of computation. Compared with the expensive cross-model interaction learning, we show that unimodal encoding with contrastive learning is a promising direction to explore for video corpus moment retrieval.

% =============================================
\begin{acks}
This research is supported by the Agency for Science, Technology and Research (A*STAR) under its AME Programmatic Fund (Project No. A18A1b0045 and A18A2b0046).
\end{acks}
% =============================================

\bibliographystyle{ACM-Reference-Format}
\bibliography{sigconf}


\begin{thebibliography}{86}

%%% the Web address.

\ifx \showCODEN    \undefined \def \showCODEN     #1{\unskip}     \fi
\ifx \showDOI      \undefined \def \showDOI       #1{#1}\fi
\ifx \showISBNx    \undefined \def \showISBNx     #1{\unskip}     \fi
\ifx \showISBNxiii \undefined \def \showISBNxiii  #1{\unskip}     \fi
\ifx \showISSN     \undefined \def \showISSN      #1{\unskip}     \fi
\ifx \showLCCN     \undefined \def \showLCCN      #1{\unskip}     \fi
\ifx \shownote     \undefined \def \shownote      #1{#1}          \fi
\ifx \showarticletitle \undefined \def \showarticletitle #1{#1}   \fi
\ifx \showURL      \undefined \def \showURL       {\relax}        \fi
% invisible to TeX
\providecommand\bibfield[2]{#2}
\providecommand\bibinfo[2]{#2}
\providecommand\natexlab[1]{#1}
\providecommand\showeprint[2][]{arXiv:#2}

\bibitem[\protect\citeauthoryear{Ba, Kiros, and Hinton}{Ba
  et~al\mbox{.}}{2016}]%
        {Ba2016LayerN}
\bibfield{author}{\bibinfo{person}{Jimmy Ba}, \bibinfo{person}{J. Kiros}, {and}
  \bibinfo{person}{Geoffrey~E. Hinton}.} \bibinfo{year}{2016}\natexlab{}.
\newblock \showarticletitle{Layer Normalization}.
\newblock \bibinfo{journal}{\emph{ArXiv}}  \bibinfo{volume}{abs/1607.06450}
  (\bibinfo{year}{2016}).
\newblock


\bibitem[\protect\citeauthoryear{Bachman, Hjelm, and Buchwalter}{Bachman
  et~al\mbox{.}}{2019}]%
        {philip2019learning}
\bibfield{author}{\bibinfo{person}{Philip Bachman}, \bibinfo{person}{R~Devon
  Hjelm}, {and} \bibinfo{person}{William Buchwalter}.}
  \bibinfo{year}{2019}\natexlab{}.
\newblock \showarticletitle{Learning Representations by Maximizing Mutual
  Information Across Views}. In \bibinfo{booktitle}{\emph{Advances in Neural
  Information Processing Systems}}, Vol.~\bibinfo{volume}{32}.
  \bibinfo{publisher}{Curran Associates, Inc.}, \bibinfo{pages}{15535--15545}.
\newblock


\bibitem[\protect\citeauthoryear{Bahdanau, Cho, and Bengio}{Bahdanau
  et~al\mbox{.}}{2015}]%
        {Bahdanau2015NeuralMT}
\bibfield{author}{\bibinfo{person}{Dzmitry Bahdanau},
  \bibinfo{person}{Kyunghyun Cho}, {and} \bibinfo{person}{Yoshua Bengio}.}
  \bibinfo{year}{2015}\natexlab{}.
\newblock \showarticletitle{Neural Machine Translation by Jointly Learning to
  Align and Translate}. In \bibinfo{booktitle}{\emph{International Conference
  on Learning Representations}}.
\newblock


\bibitem[\protect\citeauthoryear{Belghazi, Baratin, Rajeswar, Ozair, Bengio,
  Courville, and Hjelm}{Belghazi et~al\mbox{.}}{2018}]%
        {belghazi2018mine}
\bibfield{author}{\bibinfo{person}{Mohamed~Ishmael Belghazi},
  \bibinfo{person}{Aristide Baratin}, \bibinfo{person}{Sai Rajeswar},
  \bibinfo{person}{Sherjil Ozair}, \bibinfo{person}{Yoshua Bengio},
  \bibinfo{person}{Aaron Courville}, {and} \bibinfo{person}{R~Devon Hjelm}.}
  \bibinfo{year}{2018}\natexlab{}.
\newblock \showarticletitle{Mine: mutual information neural estimation}.
\newblock \bibinfo{journal}{\emph{arXiv preprint arXiv:1801.04062}}
  (\bibinfo{year}{2018}).
\newblock


\bibitem[\protect\citeauthoryear{Bell and Sejnowski}{Bell and
  Sejnowski}{1995}]%
        {bell1995information}
\bibfield{author}{\bibinfo{person}{Anthony~J Bell} {and}
  \bibinfo{person}{Terrence~J Sejnowski}.} \bibinfo{year}{1995}\natexlab{}.
\newblock \showarticletitle{An information-maximization approach to blind
  separation and blind deconvolution}.
\newblock \bibinfo{journal}{\emph{Neural computation}} \bibinfo{volume}{7},
  \bibinfo{number}{6} (\bibinfo{year}{1995}), \bibinfo{pages}{1129--1159}.
\newblock


\bibitem[\protect\citeauthoryear{Cao, Zeng, Liu, He, Wang, and Qin}{Cao
  et~al\mbox{.}}{2020}]%
        {cao2020strong}
\bibfield{author}{\bibinfo{person}{Da Cao}, \bibinfo{person}{Yawen Zeng},
  \bibinfo{person}{Meng Liu}, \bibinfo{person}{Xiangnan He},
  \bibinfo{person}{Meng Wang}, {and} \bibinfo{person}{Zheng Qin}.}
  \bibinfo{year}{2020}\natexlab{}.
\newblock \showarticletitle{STRONG: Spatio-Temporal Reinforcement Learning for
  Cross-Modal Video Moment Localization}. In
  \bibinfo{booktitle}{\emph{Proceedings of the 28th ACM International
  Conference on Multimedia}} (Seattle, WA, USA). \bibinfo{pages}{4162–4170}.
\newblock


\bibitem[\protect\citeauthoryear{Carreira and Zisserman}{Carreira and
  Zisserman}{2017}]%
        {Carreira2017QuoVA}
\bibfield{author}{\bibinfo{person}{Jo{\~a}o Carreira} {and}
  \bibinfo{person}{Andrew Zisserman}.} \bibinfo{year}{2017}\natexlab{}.
\newblock \showarticletitle{Quo Vadis, Action Recognition? A New Model and the
  Kinetics Dataset}. In \bibinfo{booktitle}{\emph{IEEE Conference on Computer
  Vision and Pattern Recognition}}. \bibinfo{pages}{4724--4733}.
\newblock


\bibitem[\protect\citeauthoryear{Chen, Chen, Ma, Jie, and Chua}{Chen
  et~al\mbox{.}}{2018}]%
        {chen2018temporally}
\bibfield{author}{\bibinfo{person}{Jingyuan Chen}, \bibinfo{person}{Xinpeng
  Chen}, \bibinfo{person}{Lin Ma}, \bibinfo{person}{Zequn Jie}, {and}
  \bibinfo{person}{Tat-Seng Chua}.} \bibinfo{year}{2018}\natexlab{}.
\newblock \showarticletitle{Temporally Grounding Natural Sentence in Video}. In
  \bibinfo{booktitle}{\emph{Proceedings of the 2018 Conference on Empirical
  Methods in Natural Language Processing}}. \bibinfo{pages}{162--171}.
\newblock


\bibitem[\protect\citeauthoryear{Chen, Lu, Tang, Xiao, Zhang, Tan, and Li}{Chen
  et~al\mbox{.}}{2020c}]%
        {chen2020rethinking}
\bibfield{author}{\bibinfo{person}{Long Chen}, \bibinfo{person}{Chujie Lu},
  \bibinfo{person}{Siliang Tang}, \bibinfo{person}{Jun Xiao},
  \bibinfo{person}{Dong Zhang}, \bibinfo{person}{Chilie Tan}, {and}
  \bibinfo{person}{Xiaolin Li}.} \bibinfo{year}{2020}\natexlab{c}.
\newblock \showarticletitle{Rethinking the Bottom-Up Framework for Query-based
  Video Localization}. In \bibinfo{booktitle}{\emph{Proceedings of the AAAI
  Conference on Artificial Intelligence}}.
\newblock


\bibitem[\protect\citeauthoryear{Chen, Jiang, Liu, and Jiang}{Chen
  et~al\mbox{.}}{2020a}]%
        {chen2020learning}
\bibfield{author}{\bibinfo{person}{Shaoxiang Chen}, \bibinfo{person}{Wenhao
  Jiang}, \bibinfo{person}{Wei Liu}, {and} \bibinfo{person}{Yu-Gang Jiang}.}
  \bibinfo{year}{2020}\natexlab{a}.
\newblock \showarticletitle{Learning Modality Interaction for Temporal Sentence
  Localization and Event Captioning in Videos}. In
  \bibinfo{booktitle}{\emph{The European Conference on Computer Vision}}.
\newblock


\bibitem[\protect\citeauthoryear{Chen and Jiang}{Chen and Jiang}{2019}]%
        {chen2019semantic}
\bibfield{author}{\bibinfo{person}{Shaoxiang Chen} {and}
  \bibinfo{person}{Yu-Gang Jiang}.} \bibinfo{year}{2019}\natexlab{}.
\newblock \showarticletitle{Semantic proposal for activity localization in
  videos via sentence query}. In \bibinfo{booktitle}{\emph{Proceedings of the
  AAAI Conference on Artificial Intelligence}}, Vol.~\bibinfo{volume}{33}.
  \bibinfo{pages}{8199--8206}.
\newblock


\bibitem[\protect\citeauthoryear{Chen, Kornblith, Norouzi, and Hinton}{Chen
  et~al\mbox{.}}{2020b}]%
        {chen2020a}
\bibfield{author}{\bibinfo{person}{Ting Chen}, \bibinfo{person}{Simon
  Kornblith}, \bibinfo{person}{Mohammad Norouzi}, {and}
  \bibinfo{person}{Geoffrey Hinton}.} \bibinfo{year}{2020}\natexlab{b}.
\newblock \showarticletitle{A Simple Framework for Contrastive Learning of
  Visual Representations}. In \bibinfo{booktitle}{\emph{Proceedings of the 37th
  International Conference on Machine Learning}}
  \emph{(\bibinfo{series}{Proceedings of Machine Learning Research},
  Vol.~\bibinfo{volume}{119})}. \bibinfo{publisher}{PMLR},
  \bibinfo{pages}{1597--1607}.
\newblock


\bibitem[\protect\citeauthoryear{Deng, Dong, Socher, Li, Li, and Fei-Fei}{Deng
  et~al\mbox{.}}{2009}]%
        {deng2009imagenet}
\bibfield{author}{\bibinfo{person}{Jia Deng}, \bibinfo{person}{Wei Dong},
  \bibinfo{person}{Richard Socher}, \bibinfo{person}{Li-Jia Li},
  \bibinfo{person}{Kai Li}, {and} \bibinfo{person}{Li Fei-Fei}.}
  \bibinfo{year}{2009}\natexlab{}.
\newblock \showarticletitle{Imagenet: A large-scale hierarchical image
  database}. In \bibinfo{booktitle}{\emph{2009 IEEE conference on computer
  vision and pattern recognition}}. Ieee, \bibinfo{pages}{248--255}.
\newblock


\bibitem[\protect\citeauthoryear{Devlin, Chang, Lee, and Toutanova}{Devlin
  et~al\mbox{.}}{2019}]%
        {devlin2019bert}
\bibfield{author}{\bibinfo{person}{Jacob Devlin}, \bibinfo{person}{Ming-Wei
  Chang}, \bibinfo{person}{Kenton Lee}, {and} \bibinfo{person}{Kristina
  Toutanova}.} \bibinfo{year}{2019}\natexlab{}.
\newblock \showarticletitle{{BERT}: Pre-training of Deep Bidirectional
  Transformers for Language Understanding}. In
  \bibinfo{booktitle}{\emph{Proceedings of the 2019 Conference of the North
  {A}merican Chapter of the Association for Computational Linguistics: Human
  Language Technologies}}. \bibinfo{publisher}{Association for Computational
  Linguistics}, \bibinfo{pages}{4171--4186}.
\newblock


\bibitem[\protect\citeauthoryear{Dong, Li, Xu, Ji, He, Yang, and Wang}{Dong
  et~al\mbox{.}}{2019}]%
        {dong2019dual}
\bibfield{author}{\bibinfo{person}{Jianfeng Dong}, \bibinfo{person}{Xirong Li},
  \bibinfo{person}{Chaoxi Xu}, \bibinfo{person}{Shouling Ji},
  \bibinfo{person}{Yuan He}, \bibinfo{person}{Gang Yang}, {and}
  \bibinfo{person}{Xun Wang}.} \bibinfo{year}{2019}\natexlab{}.
\newblock \showarticletitle{Dual encoding for zero-example video retrieval}. In
  \bibinfo{booktitle}{\emph{Proceedings of the IEEE/CVF Conference on Computer
  Vision and Pattern Recognition}}. \bibinfo{pages}{9346--9355}.
\newblock


\bibitem[\protect\citeauthoryear{Escorcia, Soldan, Sivic, Ghanem, and
  Russell}{Escorcia et~al\mbox{.}}{2019}]%
        {escorcia2019temporal}
\bibfield{author}{\bibinfo{person}{Victor Escorcia}, \bibinfo{person}{Mattia
  Soldan}, \bibinfo{person}{Josef Sivic}, \bibinfo{person}{Bernard Ghanem},
  {and} \bibinfo{person}{Bryan Russell}.} \bibinfo{year}{2019}\natexlab{}.
\newblock \showarticletitle{Temporal Localization of Moments in Video
  Collections with Natural Language}.
\newblock \bibinfo{journal}{\emph{arXiv preprint arXiv:1907.12763}}
  (\bibinfo{year}{2019}).
\newblock


\bibitem[\protect\citeauthoryear{Gabeur, Sun, Alahari, and Schmid}{Gabeur
  et~al\mbox{.}}{2020}]%
        {gabeur2020multi}
\bibfield{author}{\bibinfo{person}{Valentin Gabeur}, \bibinfo{person}{Chen
  Sun}, \bibinfo{person}{Karteek Alahari}, {and} \bibinfo{person}{Cordelia
  Schmid}.} \bibinfo{year}{2020}\natexlab{}.
\newblock \showarticletitle{Multi-modal transformer for video retrieval}. In
  \bibinfo{booktitle}{\emph{European Conference on Computer Vision (ECCV)}},
  Vol.~\bibinfo{volume}{5}. Springer.
\newblock


\bibitem[\protect\citeauthoryear{Gao, Sun, Yang, and Nevatia}{Gao
  et~al\mbox{.}}{2017}]%
        {Gao2017TALLTA}
\bibfield{author}{\bibinfo{person}{Jiyang Gao}, \bibinfo{person}{Chen Sun},
  \bibinfo{person}{Zhenheng Yang}, {and} \bibinfo{person}{Ramakant Nevatia}.}
  \bibinfo{year}{2017}\natexlab{}.
\newblock \showarticletitle{TALL: Temporal Activity Localization via Language
  Query}. In \bibinfo{booktitle}{\emph{IEEE International Conference on
  Computer Vision}}. \bibinfo{pages}{5277--5285}.
\newblock


\bibitem[\protect\citeauthoryear{Ge, Gao, Chen, and Nevatia}{Ge
  et~al\mbox{.}}{2019}]%
        {ge2019mac}
\bibfield{author}{\bibinfo{person}{Runzhou Ge}, \bibinfo{person}{Jiyang Gao},
  \bibinfo{person}{Kan Chen}, {and} \bibinfo{person}{Ram Nevatia}.}
  \bibinfo{year}{2019}\natexlab{}.
\newblock \showarticletitle{MAC: Mining Activity Concepts for Language-based
  Temporal Localization}. In \bibinfo{booktitle}{\emph{IEEE Winter Conference
  on Applications of Computer Vision}}. \bibinfo{pages}{245--253}.
\newblock


\bibitem[\protect\citeauthoryear{Ghosh, Agarwal, Parekh, and Hauptmann}{Ghosh
  et~al\mbox{.}}{2019}]%
        {ghosh2019excl}
\bibfield{author}{\bibinfo{person}{Soham Ghosh}, \bibinfo{person}{Anuva
  Agarwal}, \bibinfo{person}{Zarana Parekh}, {and} \bibinfo{person}{Alexander
  Hauptmann}.} \bibinfo{year}{2019}\natexlab{}.
\newblock \showarticletitle{{E}x{CL}: {E}xtractive {C}lip {L}ocalization
  {U}sing {N}atural {L}anguage {D}escriptions}. In
  \bibinfo{booktitle}{\emph{Proceedings of the 2019 Conference of the North
  {A}merican Chapter of the Association for Computational Linguistics: Human
  Language Technologies}}. \bibinfo{pages}{1984--1990}.
\newblock


\bibitem[\protect\citeauthoryear{Gutmann and Hyvärinen}{Gutmann and
  Hyvärinen}{2010}]%
        {michael2010noise}
\bibfield{author}{\bibinfo{person}{Michael Gutmann} {and} \bibinfo{person}{Aapo
  Hyvärinen}.} \bibinfo{year}{2010}\natexlab{}.
\newblock \showarticletitle{Noise-contrastive estimation: A new estimation
  principle for unnormalized statistical models}. In
  \bibinfo{booktitle}{\emph{Proceedings of the Thirteenth International
  Conference on Artificial Intelligence and Statistics}}
  \emph{(\bibinfo{series}{Proceedings of Machine Learning Research},
  Vol.~\bibinfo{volume}{9})}. \bibinfo{publisher}{JMLR Workshop and Conference
  Proceedings}, \bibinfo{address}{Chia Laguna Resort, Sardinia, Italy},
  \bibinfo{pages}{297--304}.
\newblock


\bibitem[\protect\citeauthoryear{Hadsell, Chopra, and LeCun}{Hadsell
  et~al\mbox{.}}{2006}]%
        {hadsell2006dimensionality}
\bibfield{author}{\bibinfo{person}{Raia Hadsell}, \bibinfo{person}{Sumit
  Chopra}, {and} \bibinfo{person}{Yann LeCun}.}
  \bibinfo{year}{2006}\natexlab{}.
\newblock \showarticletitle{Dimensionality reduction by learning an invariant
  mapping}. In \bibinfo{booktitle}{\emph{2006 IEEE Computer Society Conference
  on Computer Vision and Pattern Recognition (CVPR'06)}},
  Vol.~\bibinfo{volume}{2}. IEEE, \bibinfo{pages}{1735--1742}.
\newblock


\bibitem[\protect\citeauthoryear{Hahn, Kadav, Rehg, and Graf}{Hahn
  et~al\mbox{.}}{2020}]%
        {hahn2020tripping}
\bibfield{author}{\bibinfo{person}{Meera Hahn}, \bibinfo{person}{Asim Kadav},
  \bibinfo{person}{James~M Rehg}, {and} \bibinfo{person}{Hans~Peter Graf}.}
  \bibinfo{year}{2020}\natexlab{}.
\newblock \showarticletitle{Tripping through time: Efficient localization of
  activities in videos}. In \bibinfo{booktitle}{\emph{The British Machine
  Vision Conference}}.
\newblock


\bibitem[\protect\citeauthoryear{He, Zhao, Huang, Li, Liu, and Wen}{He
  et~al\mbox{.}}{2019}]%
        {he2019Readwa}
\bibfield{author}{\bibinfo{person}{Dongliang He}, \bibinfo{person}{Xiang Zhao},
  \bibinfo{person}{Jizhou Huang}, \bibinfo{person}{Fu Li},
  \bibinfo{person}{Xiao Liu}, {and} \bibinfo{person}{Shilei Wen}.}
  \bibinfo{year}{2019}\natexlab{}.
\newblock \showarticletitle{Read, watch, and move: Reinforcement learning for
  temporally grounding natural language descriptions in videos}. In
  \bibinfo{booktitle}{\emph{Proceedings of the AAAI Conference on Artificial
  Intelligence}}, Vol.~\bibinfo{volume}{33}. \bibinfo{pages}{8393--8400}.
\newblock


\bibitem[\protect\citeauthoryear{He, Fan, Wu, Xie, and Girshick}{He
  et~al\mbox{.}}{2020}]%
        {he2020momentum}
\bibfield{author}{\bibinfo{person}{Kaiming He}, \bibinfo{person}{Haoqi Fan},
  \bibinfo{person}{Yuxin Wu}, \bibinfo{person}{Saining Xie}, {and}
  \bibinfo{person}{Ross Girshick}.} \bibinfo{year}{2020}\natexlab{}.
\newblock \showarticletitle{Momentum contrast for unsupervised visual
  representation learning}. In \bibinfo{booktitle}{\emph{Proceedings of the
  IEEE/CVF Conference on Computer Vision and Pattern Recognition}}.
  \bibinfo{pages}{9729--9738}.
\newblock


\bibitem[\protect\citeauthoryear{He, Zhang, Ren, and Sun}{He
  et~al\mbox{.}}{2016}]%
        {he2016resnet}
\bibfield{author}{\bibinfo{person}{Kaiming He}, \bibinfo{person}{Xiangyu
  Zhang}, \bibinfo{person}{Shaoqing Ren}, {and} \bibinfo{person}{Jian Sun}.}
  \bibinfo{year}{2016}\natexlab{}.
\newblock \showarticletitle{Deep residual learning for image recognition}. In
  \bibinfo{booktitle}{\emph{Proceedings of the IEEE conference on computer
  vision and pattern recognition}}. \bibinfo{pages}{770--778}.
\newblock


\bibitem[\protect\citeauthoryear{{Heilbron}, {Escorcia}, {Ghanem}, and
  {Niebles}}{{Heilbron} et~al\mbox{.}}{2015}]%
        {heilbron2015activitynet}
\bibfield{author}{\bibinfo{person}{F.~C. {Heilbron}}, \bibinfo{person}{V.
  {Escorcia}}, \bibinfo{person}{B. {Ghanem}}, {and} \bibinfo{person}{J.~C.
  {Niebles}}.} \bibinfo{year}{2015}\natexlab{}.
\newblock \showarticletitle{ActivityNet: A large-scale video benchmark for
  human activity understanding}. In \bibinfo{booktitle}{\emph{IEEE Conference
  on Computer Vision and Pattern Recognition}}. \bibinfo{pages}{961--970}.
\newblock


\bibitem[\protect\citeauthoryear{Hendricks, Wang, Shechtman, Sivic, Darrell,
  and Russell}{Hendricks et~al\mbox{.}}{2018}]%
        {hendricks2018localizingM}
\bibfield{author}{\bibinfo{person}{Lisa~Anne Hendricks},
  \bibinfo{person}{Oliver Wang}, \bibinfo{person}{Eli Shechtman},
  \bibinfo{person}{Josef Sivic}, \bibinfo{person}{Trevor Darrell}, {and}
  \bibinfo{person}{Bryan Russell}.} \bibinfo{year}{2018}\natexlab{}.
\newblock \showarticletitle{Localizing Moments in Video with Temporal
  Language}. In \bibinfo{booktitle}{\emph{Proceedings of the 2018 Conference on
  Empirical Methods in Natural Language Processing}}.
  \bibinfo{pages}{1380--1390}.
\newblock


\bibitem[\protect\citeauthoryear{Hendricks, Wang, Shechtman, Sivic, Darrell,
  and Russell}{Hendricks et~al\mbox{.}}{2017a}]%
        {Hendricks2017LocalizingMI}
\bibfield{author}{\bibinfo{person}{Lisa~Anne Hendricks},
  \bibinfo{person}{Oliver Wang}, \bibinfo{person}{Eli Shechtman},
  \bibinfo{person}{Josef Sivic}, \bibinfo{person}{Trevor Darrell}, {and}
  \bibinfo{person}{Bryan~C. Russell}.} \bibinfo{year}{2017}\natexlab{a}.
\newblock \showarticletitle{Localizing Moments in Video with Natural Language}.
  In \bibinfo{booktitle}{\emph{2017 IEEE International Conference on Computer
  Vision (ICCV)}}. \bibinfo{pages}{5804--5813}.
\newblock


\bibitem[\protect\citeauthoryear{Hendricks, Wang, Shechtman, Sivic, Darrell,
  and Russell}{Hendricks et~al\mbox{.}}{2017b}]%
        {hendricks2017localizing}
\bibfield{author}{\bibinfo{person}{Lisa~Anne Hendricks},
  \bibinfo{person}{Oliver Wang}, \bibinfo{person}{Eli Shechtman},
  \bibinfo{person}{Josef Sivic}, \bibinfo{person}{Trevor Darrell}, {and}
  \bibinfo{person}{Bryan~C. Russell}.} \bibinfo{year}{2017}\natexlab{b}.
\newblock \showarticletitle{Localizing Moments in Video with Natural Language}.
  In \bibinfo{booktitle}{\emph{2017 IEEE International Conference on Computer
  Vision}}. \bibinfo{pages}{5804--5813}.
\newblock


\bibitem[\protect\citeauthoryear{Hjelm, Fedorov, Lavoie-Marchildon, Grewal,
  Bachman, Trischler, and Bengio}{Hjelm et~al\mbox{.}}{2019}]%
        {hjelm2018learning}
\bibfield{author}{\bibinfo{person}{R~Devon Hjelm}, \bibinfo{person}{Alex
  Fedorov}, \bibinfo{person}{Samuel Lavoie-Marchildon}, \bibinfo{person}{Karan
  Grewal}, \bibinfo{person}{Phil Bachman}, \bibinfo{person}{Adam Trischler},
  {and} \bibinfo{person}{Yoshua Bengio}.} \bibinfo{year}{2019}\natexlab{}.
\newblock \showarticletitle{Learning deep representations by mutual information
  estimation and maximization}. In \bibinfo{booktitle}{\emph{International
  Conference on Learning Representations}}.
\newblock


\bibitem[\protect\citeauthoryear{Huang, Zhu, Shen, and Chen}{Huang
  et~al\mbox{.}}{2018}]%
        {huang2018fusionnet}
\bibfield{author}{\bibinfo{person}{Hsin-Yuan Huang}, \bibinfo{person}{Chenguang
  Zhu}, \bibinfo{person}{Yelong Shen}, {and} \bibinfo{person}{Weizhu Chen}.}
  \bibinfo{year}{2018}\natexlab{}.
\newblock \showarticletitle{FusionNet: Fusing via Fully-aware Attention with
  Application to Machine Comprehension}. In
  \bibinfo{booktitle}{\emph{International Conference on Learning
  Representations}}.
\newblock


\bibitem[\protect\citeauthoryear{Hyv{\"a}rinen and Oja}{Hyv{\"a}rinen and
  Oja}{2000}]%
        {hyvarinen2000independent}
\bibfield{author}{\bibinfo{person}{Aapo Hyv{\"a}rinen} {and}
  \bibinfo{person}{Erkki Oja}.} \bibinfo{year}{2000}\natexlab{}.
\newblock \showarticletitle{Independent component analysis: algorithms and
  applications}.
\newblock \bibinfo{journal}{\emph{Neural networks}} \bibinfo{volume}{13},
  \bibinfo{number}{4-5} (\bibinfo{year}{2000}), \bibinfo{pages}{411--430}.
\newblock


\bibitem[\protect\citeauthoryear{Jozefowicz, Vinyals, Schuster, Shazeer, and
  Wu}{Jozefowicz et~al\mbox{.}}{2016}]%
        {jozefowicz2016exploring}
\bibfield{author}{\bibinfo{person}{Rafal Jozefowicz}, \bibinfo{person}{Oriol
  Vinyals}, \bibinfo{person}{Mike Schuster}, \bibinfo{person}{Noam Shazeer},
  {and} \bibinfo{person}{Yonghui Wu}.} \bibinfo{year}{2016}\natexlab{}.
\newblock \showarticletitle{Exploring the limits of language modeling}.
\newblock \bibinfo{journal}{\emph{arXiv preprint arXiv:1602.02410}}
  (\bibinfo{year}{2016}).
\newblock


\bibitem[\protect\citeauthoryear{Kay, Carreira, Simonyan, Zhang, Hillier,
  Vijayanarasimhan, Viola, Green, Back, Natsev, et~al\mbox{.}}{Kay
  et~al\mbox{.}}{2017}]%
        {kay2017kinetics}
\bibfield{author}{\bibinfo{person}{Will Kay}, \bibinfo{person}{Joao Carreira},
  \bibinfo{person}{Karen Simonyan}, \bibinfo{person}{Brian Zhang},
  \bibinfo{person}{Chloe Hillier}, \bibinfo{person}{Sudheendra
  Vijayanarasimhan}, \bibinfo{person}{Fabio Viola}, \bibinfo{person}{Tim
  Green}, \bibinfo{person}{Trevor Back}, \bibinfo{person}{Paul Natsev},
  {et~al\mbox{.}}} \bibinfo{year}{2017}\natexlab{}.
\newblock \showarticletitle{The kinetics human action video dataset}.
\newblock \bibinfo{journal}{\emph{arXiv preprint arXiv:1705.06950}}
  (\bibinfo{year}{2017}).
\newblock


\bibitem[\protect\citeauthoryear{{Krishna}, {Hata}, {Ren}, {Fei-Fei}, and
  {Niebles}}{{Krishna} et~al\mbox{.}}{2017}]%
        {krishna2017dense}
\bibfield{author}{\bibinfo{person}{R. {Krishna}}, \bibinfo{person}{K. {Hata}},
  \bibinfo{person}{F. {Ren}}, \bibinfo{person}{L. {Fei-Fei}}, {and}
  \bibinfo{person}{J.~C. {Niebles}}.} \bibinfo{year}{2017}\natexlab{}.
\newblock \showarticletitle{Dense-Captioning Events in Videos}. In
  \bibinfo{booktitle}{\emph{IEEE International Conference on Computer Vision}}.
  \bibinfo{pages}{706--715}.
\newblock


\bibitem[\protect\citeauthoryear{Lei, Yu, Berg, and Bansal}{Lei
  et~al\mbox{.}}{2020}]%
        {lei2020tvr}
\bibfield{author}{\bibinfo{person}{Jie Lei}, \bibinfo{person}{Licheng Yu},
  \bibinfo{person}{Tamara~L Berg}, {and} \bibinfo{person}{Mohit Bansal}.}
  \bibinfo{year}{2020}\natexlab{}.
\newblock \showarticletitle{TVR: A Large-Scale Dataset for Video-Subtitle
  Moment Retrieval}. In \bibinfo{booktitle}{\emph{The European Conference on
  Computer Vision}}.
\newblock


\bibitem[\protect\citeauthoryear{Li, Chen, Cheng, Gan, Yu, and Liu}{Li
  et~al\mbox{.}}{2020a}]%
        {li2020hero}
\bibfield{author}{\bibinfo{person}{Linjie Li}, \bibinfo{person}{Yen-Chun Chen},
  \bibinfo{person}{Yu Cheng}, \bibinfo{person}{Zhe Gan},
  \bibinfo{person}{Licheng Yu}, {and} \bibinfo{person}{Jingjing Liu}.}
  \bibinfo{year}{2020}\natexlab{a}.
\newblock \showarticletitle{{HERO}: Hierarchical Encoder for {V}ideo+{L}anguage
  Omni-representation Pre-training}. In \bibinfo{booktitle}{\emph{Proceedings
  of the 2020 Conference on Empirical Methods in Natural Language Processing
  (EMNLP)}}. \bibinfo{publisher}{Association for Computational Linguistics},
  \bibinfo{address}{Online}, \bibinfo{pages}{2046--2065}.
\newblock


\bibitem[\protect\citeauthoryear{Li, Zhou, Xu, Ji, and Yang}{Li
  et~al\mbox{.}}{2020b}]%
        {li2020sea}
\bibfield{author}{\bibinfo{person}{Xirong Li}, \bibinfo{person}{Fangming Zhou},
  \bibinfo{person}{Chaoxi Xu}, \bibinfo{person}{Jiaqi Ji}, {and}
  \bibinfo{person}{Gang Yang}.} \bibinfo{year}{2020}\natexlab{b}.
\newblock \showarticletitle{SEA: Sentence Encoder Assembly for Video Retrieval
  by Textual Queries}.
\newblock \bibinfo{journal}{\emph{IEEE Transactions on Multimedia}}
  (\bibinfo{year}{2020}).
\newblock


\bibitem[\protect\citeauthoryear{Lin, Zhao, Su, Wang, and Yang}{Lin
  et~al\mbox{.}}{2018}]%
        {lin2018bsn}
\bibfield{author}{\bibinfo{person}{Tianwei Lin}, \bibinfo{person}{Xu Zhao},
  \bibinfo{person}{Haisheng Su}, \bibinfo{person}{Chongjing Wang}, {and}
  \bibinfo{person}{Ming Yang}.} \bibinfo{year}{2018}\natexlab{}.
\newblock \showarticletitle{Bsn: Boundary sensitive network for temporal action
  proposal generation}. In \bibinfo{booktitle}{\emph{Proceedings of the
  European Conference on Computer Vision}}. \bibinfo{pages}{3--19}.
\newblock


\bibitem[\protect\citeauthoryear{Lin, Zhao, Zhang, Wang, and Liu}{Lin
  et~al\mbox{.}}{2020}]%
        {Lin2020WeaklySupervisedVM}
\bibfield{author}{\bibinfo{person}{Zhijie Lin}, \bibinfo{person}{Zhou Zhao},
  \bibinfo{person}{Zhu Zhang}, \bibinfo{person}{Qi Wang}, {and}
  \bibinfo{person}{Huasheng Liu}.} \bibinfo{year}{2020}\natexlab{}.
\newblock \showarticletitle{Weakly-Supervised Video Moment Retrieval via
  Semantic Completion Network}. In \bibinfo{booktitle}{\emph{Proceedings of the
  AAAI Conference on Artificial Intelligence}}.
\newblock


\bibitem[\protect\citeauthoryear{Liu, Wang, Nie, Tian, Chen, and Chua}{Liu
  et~al\mbox{.}}{2018}]%
        {Liu2018CML}
\bibfield{author}{\bibinfo{person}{Meng Liu}, \bibinfo{person}{Xiang Wang},
  \bibinfo{person}{Liqiang Nie}, \bibinfo{person}{Qi Tian},
  \bibinfo{person}{Baoquan Chen}, {and} \bibinfo{person}{Tat-Seng Chua}.}
  \bibinfo{year}{2018}\natexlab{}.
\newblock \showarticletitle{Cross-modal Moment Localization in Videos}. In
  \bibinfo{booktitle}{\emph{Proceedings of the 26th ACM International
  Conference on Multimedia}}. \bibinfo{pages}{843--851}.
\newblock
\showISBNx{978-1-4503-5665-7}


\bibitem[\protect\citeauthoryear{Liu, Ott, Goyal, Du, Joshi, Chen, Levy, Lewis,
  Zettlemoyer, and Stoyanov}{Liu et~al\mbox{.}}{2019}]%
        {liu2019roberta}
\bibfield{author}{\bibinfo{person}{Yinhan Liu}, \bibinfo{person}{Myle Ott},
  \bibinfo{person}{Naman Goyal}, \bibinfo{person}{Jingfei Du},
  \bibinfo{person}{Mandar Joshi}, \bibinfo{person}{Danqi Chen},
  \bibinfo{person}{Omer Levy}, \bibinfo{person}{Mike Lewis},
  \bibinfo{person}{Luke Zettlemoyer}, {and} \bibinfo{person}{Veselin
  Stoyanov}.} \bibinfo{year}{2019}\natexlab{}.
\newblock \showarticletitle{Roberta: A robustly optimized bert pretraining
  approach}.
\newblock \bibinfo{journal}{\emph{arXiv preprint arXiv:1907.11692}}
  (\bibinfo{year}{2019}).
\newblock


\bibitem[\protect\citeauthoryear{Lu, Chen, Tan, Li, and Xiao}{Lu
  et~al\mbox{.}}{2019b}]%
        {lu2019debug}
\bibfield{author}{\bibinfo{person}{Chujie Lu}, \bibinfo{person}{Long Chen},
  \bibinfo{person}{Chilie Tan}, \bibinfo{person}{Xiaolin Li}, {and}
  \bibinfo{person}{Jun Xiao}.} \bibinfo{year}{2019}\natexlab{b}.
\newblock \showarticletitle{{DEBUG}: A Dense Bottom-Up Grounding Approach for
  Natural Language Video Localization}. In
  \bibinfo{booktitle}{\emph{Proceedings of the 2019 Conference on Empirical
  Methods in Natural Language Processing and the 9th International Joint
  Conference on Natural Language Processing}}. \bibinfo{pages}{5147--5156}.
\newblock


\bibitem[\protect\citeauthoryear{Lu, Batra, Parikh, and Lee}{Lu
  et~al\mbox{.}}{2019a}]%
        {lu2019vilbert}
\bibfield{author}{\bibinfo{person}{Jiasen Lu}, \bibinfo{person}{Dhruv Batra},
  \bibinfo{person}{Devi Parikh}, {and} \bibinfo{person}{Stefan Lee}.}
  \bibinfo{year}{2019}\natexlab{a}.
\newblock \showarticletitle{ViLBERT: Pretraining Task-Agnostic Visiolinguistic
  Representations for Vision-and-Language Tasks}. In
  \bibinfo{booktitle}{\emph{Advances in Neural Information Processing
  Systems}}. \bibinfo{pages}{13--23}.
\newblock


\bibitem[\protect\citeauthoryear{Luo, Ji, Shi, Huang, Duan, Li, Chen, and
  Zhou}{Luo et~al\mbox{.}}{2020}]%
        {luo2020univilm}
\bibfield{author}{\bibinfo{person}{Huaishao Luo}, \bibinfo{person}{Lei Ji},
  \bibinfo{person}{Botian Shi}, \bibinfo{person}{Haoyang Huang},
  \bibinfo{person}{Nan Duan}, \bibinfo{person}{Tianrui Li},
  \bibinfo{person}{Xilin Chen}, {and} \bibinfo{person}{Ming Zhou}.}
  \bibinfo{year}{2020}\natexlab{}.
\newblock \showarticletitle{Univilm: A unified video and language pre-training
  model for multimodal understanding and generation}.
\newblock \bibinfo{journal}{\emph{arXiv preprint arXiv:2002.06353}}
  (\bibinfo{year}{2020}).
\newblock


\bibitem[\protect\citeauthoryear{Miech, Alayrac, Smaira, Laptev, Sivic, and
  Zisserman}{Miech et~al\mbox{.}}{2020}]%
        {miech2020end}
\bibfield{author}{\bibinfo{person}{Antoine Miech},
  \bibinfo{person}{Jean-Baptiste Alayrac}, \bibinfo{person}{Lucas Smaira},
  \bibinfo{person}{Ivan Laptev}, \bibinfo{person}{Josef Sivic}, {and}
  \bibinfo{person}{Andrew Zisserman}.} \bibinfo{year}{2020}\natexlab{}.
\newblock \showarticletitle{End-to-end learning of visual representations from
  uncurated instructional videos}. In \bibinfo{booktitle}{\emph{Proceedings of
  the IEEE/CVF Conference on Computer Vision and Pattern Recognition}}.
  \bibinfo{pages}{9879--9889}.
\newblock


\bibitem[\protect\citeauthoryear{Miech, Laptev, and Sivic}{Miech
  et~al\mbox{.}}{2018}]%
        {miech2018learning}
\bibfield{author}{\bibinfo{person}{Antoine Miech}, \bibinfo{person}{Ivan
  Laptev}, {and} \bibinfo{person}{Josef Sivic}.}
  \bibinfo{year}{2018}\natexlab{}.
\newblock \showarticletitle{Learning a text-video embedding from incomplete and
  heterogeneous data}.
\newblock \bibinfo{journal}{\emph{arXiv preprint arXiv:1804.02516}}
  (\bibinfo{year}{2018}).
\newblock


\bibitem[\protect\citeauthoryear{Misra and Maaten}{Misra and Maaten}{2020}]%
        {misra2020self}
\bibfield{author}{\bibinfo{person}{Ishan Misra} {and} \bibinfo{person}{Laurens
  van~der Maaten}.} \bibinfo{year}{2020}\natexlab{}.
\newblock \showarticletitle{Self-supervised learning of pretext-invariant
  representations}. In \bibinfo{booktitle}{\emph{Proceedings of the IEEE/CVF
  Conference on Computer Vision and Pattern Recognition}}.
  \bibinfo{pages}{6707--6717}.
\newblock


\bibitem[\protect\citeauthoryear{Mithun, Li, Metze, and Roy-Chowdhury}{Mithun
  et~al\mbox{.}}{2018}]%
        {mithun2018learning}
\bibfield{author}{\bibinfo{person}{Niluthpol~Chowdhury Mithun},
  \bibinfo{person}{Juncheng Li}, \bibinfo{person}{Florian Metze}, {and}
  \bibinfo{person}{Amit~K Roy-Chowdhury}.} \bibinfo{year}{2018}\natexlab{}.
\newblock \showarticletitle{Learning joint embedding with multimodal cues for
  cross-modal video-text retrieval}. In \bibinfo{booktitle}{\emph{Proceedings
  of the 2018 ACM on International Conference on Multimedia Retrieval}}.
  \bibinfo{pages}{19--27}.
\newblock


\bibitem[\protect\citeauthoryear{Mithun, Paul, and Roy-Chowdhury}{Mithun
  et~al\mbox{.}}{2019}]%
        {mithun2019weakly}
\bibfield{author}{\bibinfo{person}{Niluthpol~Chowdhury Mithun},
  \bibinfo{person}{Sujoy Paul}, {and} \bibinfo{person}{Amit~K. Roy-Chowdhury}.}
  \bibinfo{year}{2019}\natexlab{}.
\newblock \showarticletitle{Weakly Supervised Video Moment Retrieval From Text
  Queries}. In \bibinfo{booktitle}{\emph{Proceedings of the IEEE Conference on
  Computer Vision and Pattern Recognition}}. \bibinfo{pages}{11592--11601}.
\newblock


\bibitem[\protect\citeauthoryear{Mun, Cho, and Han}{Mun et~al\mbox{.}}{2020}]%
        {mun2020local}
\bibfield{author}{\bibinfo{person}{Jonghwan Mun}, \bibinfo{person}{Minsu Cho},
  {and} \bibinfo{person}{Bohyung Han}.} \bibinfo{year}{2020}\natexlab{}.
\newblock \showarticletitle{Local-Global Video-Text Interactions for Temporal
  Grounding}. In \bibinfo{booktitle}{\emph{Proceedings of the IEEE/CVF
  Conference on Computer Vision and Pattern Recognition}}.
  \bibinfo{pages}{10810--10819}.
\newblock


\bibitem[\protect\citeauthoryear{Oord, Li, and Vinyals}{Oord
  et~al\mbox{.}}{2018}]%
        {oord2018representation}
\bibfield{author}{\bibinfo{person}{Aaron van~den Oord}, \bibinfo{person}{Yazhe
  Li}, {and} \bibinfo{person}{Oriol Vinyals}.} \bibinfo{year}{2018}\natexlab{}.
\newblock \showarticletitle{Representation learning with contrastive predictive
  coding}.
\newblock \bibinfo{journal}{\emph{arXiv preprint arXiv:1807.03748}}
  (\bibinfo{year}{2018}).
\newblock


\bibitem[\protect\citeauthoryear{Pan, Mei, Yao, Li, and Rui}{Pan
  et~al\mbox{.}}{2016}]%
        {pan2016jointly}
\bibfield{author}{\bibinfo{person}{Yingwei Pan}, \bibinfo{person}{Tao Mei},
  \bibinfo{person}{Ting Yao}, \bibinfo{person}{Houqiang Li}, {and}
  \bibinfo{person}{Yong Rui}.} \bibinfo{year}{2016}\natexlab{}.
\newblock \showarticletitle{Jointly modeling embedding and translation to
  bridge video and language}. In \bibinfo{booktitle}{\emph{Proceedings of the
  IEEE conference on computer vision and pattern recognition}}.
  \bibinfo{pages}{4594--4602}.
\newblock


\bibitem[\protect\citeauthoryear{Pennington, Socher, and Manning}{Pennington
  et~al\mbox{.}}{2014}]%
        {pennington2014glove}
\bibfield{author}{\bibinfo{person}{Jeffrey Pennington},
  \bibinfo{person}{Richard Socher}, {and} \bibinfo{person}{Christopher
  Manning}.} \bibinfo{year}{2014}\natexlab{}.
\newblock \showarticletitle{{G}love: Global Vectors for Word Representation}.
  In \bibinfo{booktitle}{\emph{Proceedings of the 2014 Conference on Empirical
  Methods in Natural Language Processing}}. \bibinfo{pages}{1532--1543}.
\newblock


\bibitem[\protect\citeauthoryear{Rodriguez, Marrese-Taylor, Saleh, Li, and
  Gould}{Rodriguez et~al\mbox{.}}{2020}]%
        {rodriguez2020proposal}
\bibfield{author}{\bibinfo{person}{Cristian Rodriguez}, \bibinfo{person}{Edison
  Marrese-Taylor}, \bibinfo{person}{Fatemeh~Sadat Saleh},
  \bibinfo{person}{Hongdong Li}, {and} \bibinfo{person}{Stephen Gould}.}
  \bibinfo{year}{2020}\natexlab{}.
\newblock \showarticletitle{Proposal-free Temporal Moment Localization of a
  Natural-Language Query in Video using Guided Attention}. In
  \bibinfo{booktitle}{\emph{The IEEE Winter Conference on Applications of
  Computer Vision}}.
\newblock


\bibitem[\protect\citeauthoryear{Seo, Kembhavi, Farhadi, and Hajishirzi}{Seo
  et~al\mbox{.}}{2017}]%
        {Seo2017BidirectionalAF}
\bibfield{author}{\bibinfo{person}{Minjoon Seo}, \bibinfo{person}{Aniruddha
  Kembhavi}, \bibinfo{person}{Ali Farhadi}, {and} \bibinfo{person}{Hannaneh
  Hajishirzi}.} \bibinfo{year}{2017}\natexlab{}.
\newblock \showarticletitle{Bidirectional Attention Flow for Machine
  Comprehension}. In \bibinfo{booktitle}{\emph{International Conference on
  Learning Representations}}.
\newblock


\bibitem[\protect\citeauthoryear{Shao, Xiong, Zhao, Huang, Qiao, and Lin}{Shao
  et~al\mbox{.}}{2018}]%
        {shao2018find}
\bibfield{author}{\bibinfo{person}{Dian Shao}, \bibinfo{person}{Yu Xiong},
  \bibinfo{person}{Yue Zhao}, \bibinfo{person}{Qingqiu Huang},
  \bibinfo{person}{Yu Qiao}, {and} \bibinfo{person}{Dahua Lin}.}
  \bibinfo{year}{2018}\natexlab{}.
\newblock \showarticletitle{Find and focus: Retrieve and localize video events
  with natural language queries}. In \bibinfo{booktitle}{\emph{Proceedings of
  the European Conference on Computer Vision (ECCV)}}.
  \bibinfo{pages}{200--216}.
\newblock


\bibitem[\protect\citeauthoryear{Sun, Baradel, Murphy, and Schmid}{Sun
  et~al\mbox{.}}{2019}]%
        {sun2019cbt}
\bibfield{author}{\bibinfo{person}{Chen Sun}, \bibinfo{person}{Fabien Baradel},
  \bibinfo{person}{Kevin Murphy}, {and} \bibinfo{person}{Cordelia Schmid}.}
  \bibinfo{year}{2019}\natexlab{}.
\newblock \showarticletitle{Learning video representations using contrastive
  bidirectional transformer}.
\newblock \bibinfo{journal}{\emph{arXiv preprint arXiv:1906.05743}}
  (\bibinfo{year}{2019}).
\newblock


\bibitem[\protect\citeauthoryear{Sun, Hoffman, Verma, and Tang}{Sun
  et~al\mbox{.}}{2020}]%
        {Sun2020InfoGraph}
\bibfield{author}{\bibinfo{person}{Fan-Yun Sun}, \bibinfo{person}{Jordan
  Hoffman}, \bibinfo{person}{Vikas Verma}, {and} \bibinfo{person}{Jian Tang}.}
  \bibinfo{year}{2020}\natexlab{}.
\newblock \showarticletitle{InfoGraph: Unsupervised and Semi-supervised
  Graph-Level Representation Learning via Mutual Information Maximization}. In
  \bibinfo{booktitle}{\emph{International Conference on Learning
  Representations}}.
\newblock


\bibitem[\protect\citeauthoryear{Tan and Bansal}{Tan and Bansal}{2019}]%
        {tan2019lxmert}
\bibfield{author}{\bibinfo{person}{Hao Tan} {and} \bibinfo{person}{Mohit
  Bansal}.} \bibinfo{year}{2019}\natexlab{}.
\newblock \showarticletitle{{LXMERT}: Learning Cross-Modality Encoder
  Representations from Transformers}. In \bibinfo{booktitle}{\emph{Proceedings
  of the 2019 Conference on Empirical Methods in Natural Language Processing
  and the 9th International Joint Conference on Natural Language Processing}}.
  \bibinfo{pages}{5103--5114}.
\newblock


\bibitem[\protect\citeauthoryear{Vaswani, Shazeer, Parmar, Uszkoreit, Jones,
  Gomez, Kaiser, and Polosukhin}{Vaswani et~al\mbox{.}}{2017}]%
        {vaswani2017attention}
\bibfield{author}{\bibinfo{person}{Ashish Vaswani}, \bibinfo{person}{Noam
  Shazeer}, \bibinfo{person}{Niki Parmar}, \bibinfo{person}{Jakob Uszkoreit},
  \bibinfo{person}{Llion Jones}, \bibinfo{person}{Aidan~N Gomez},
  \bibinfo{person}{\L~ukasz Kaiser}, {and} \bibinfo{person}{Illia Polosukhin}.}
  \bibinfo{year}{2017}\natexlab{}.
\newblock \showarticletitle{Attention is All you Need}. In
  \bibinfo{booktitle}{\emph{Advances in Neural Information Processing
  Systems}}. \bibinfo{pages}{5998--6008}.
\newblock


\bibitem[\protect\citeauthoryear{Veličković, Fedus, Hamilton, Liò, Bengio,
  and Hjelm}{Veličković et~al\mbox{.}}{2019}]%
        {velickovic2018deep}
\bibfield{author}{\bibinfo{person}{Petar Veličković},
  \bibinfo{person}{William Fedus}, \bibinfo{person}{William~L. Hamilton},
  \bibinfo{person}{Pietro Liò}, \bibinfo{person}{Yoshua Bengio}, {and}
  \bibinfo{person}{R~Devon Hjelm}.} \bibinfo{year}{2019}\natexlab{}.
\newblock \showarticletitle{Deep Graph Infomax}. In
  \bibinfo{booktitle}{\emph{International Conference on Learning
  Representations}}.
\newblock


\bibitem[\protect\citeauthoryear{Venugopalan, Rohrbach, Donahue, Mooney,
  Darrell, and Saenko}{Venugopalan et~al\mbox{.}}{2015}]%
        {venugopalan2015sequence}
\bibfield{author}{\bibinfo{person}{Subhashini Venugopalan},
  \bibinfo{person}{Marcus Rohrbach}, \bibinfo{person}{Jeffrey Donahue},
  \bibinfo{person}{Raymond Mooney}, \bibinfo{person}{Trevor Darrell}, {and}
  \bibinfo{person}{Kate Saenko}.} \bibinfo{year}{2015}\natexlab{}.
\newblock \showarticletitle{Sequence to sequence-video to text}. In
  \bibinfo{booktitle}{\emph{Proceedings of the IEEE international conference on
  computer vision}}. \bibinfo{pages}{4534--4542}.
\newblock


\bibitem[\protect\citeauthoryear{Wang, Zha, Chen, Xiong, and Luo}{Wang
  et~al\mbox{.}}{2020b}]%
        {wang2020dual}
\bibfield{author}{\bibinfo{person}{Hao Wang}, \bibinfo{person}{Zheng-Jun Zha},
  \bibinfo{person}{Xuejin Chen}, \bibinfo{person}{Zhiwei Xiong}, {and}
  \bibinfo{person}{Jiebo Luo}.} \bibinfo{year}{2020}\natexlab{b}.
\newblock \showarticletitle{Dual Path Interaction Network for Video Moment
  Localization}. In \bibinfo{booktitle}{\emph{Proceedings of the 28th ACM
  International Conference on Multimedia}} (Seattle, WA, USA).
  \bibinfo{pages}{4116–4124}.
\newblock


\bibitem[\protect\citeauthoryear{Wang, Ma, and Jiang}{Wang
  et~al\mbox{.}}{2020a}]%
        {Wang2020TemporallyGL}
\bibfield{author}{\bibinfo{person}{Jingwen Wang}, \bibinfo{person}{Lin Ma},
  {and} \bibinfo{person}{Wenhao Jiang}.} \bibinfo{year}{2020}\natexlab{a}.
\newblock \showarticletitle{Temporally Grounding Language Queries in Videos by
  Contextual Boundary-aware Prediction}. In
  \bibinfo{booktitle}{\emph{Proceedings of the AAAI Conference on Artificial
  Intelligence}}.
\newblock


\bibitem[\protect\citeauthoryear{Wang, Huang, and Wang}{Wang
  et~al\mbox{.}}{2019}]%
        {Wang2019LanguageDrivenTA}
\bibfield{author}{\bibinfo{person}{Weining Wang}, \bibinfo{person}{Yan Huang},
  {and} \bibinfo{person}{Liang Wang}.} \bibinfo{year}{2019}\natexlab{}.
\newblock \showarticletitle{Language-Driven Temporal Activity Localization: A
  Semantic Matching Reinforcement Learning Model}. In
  \bibinfo{booktitle}{\emph{Proceedings of the IEEE Conference on Computer
  Vision and Pattern Recognition}}. \bibinfo{pages}{334--343}.
\newblock


\bibitem[\protect\citeauthoryear{Wang, Yang, Wei, Chang, and Zhou}{Wang
  et~al\mbox{.}}{2017}]%
        {wang2017gated}
\bibfield{author}{\bibinfo{person}{Wenhui Wang}, \bibinfo{person}{Nan Yang},
  \bibinfo{person}{Furu Wei}, \bibinfo{person}{Baobao Chang}, {and}
  \bibinfo{person}{Ming Zhou}.} \bibinfo{year}{2017}\natexlab{}.
\newblock \showarticletitle{Gated Self-Matching Networks for Reading
  Comprehension and Question Answering}. In
  \bibinfo{booktitle}{\emph{Proceedings of the 55th Annual Meeting of the
  Association for Computational Linguistics}}. \bibinfo{publisher}{Association
  for Computational Linguistics}, \bibinfo{pages}{189--198}.
\newblock


\bibitem[\protect\citeauthoryear{Wu, Li, Han, and Lin}{Wu
  et~al\mbox{.}}{2020a}]%
        {wu2020reinforcement}
\bibfield{author}{\bibinfo{person}{Jie Wu}, \bibinfo{person}{Guanbin Li},
  \bibinfo{person}{Xiaoguang Han}, {and} \bibinfo{person}{Liang Lin}.}
  \bibinfo{year}{2020}\natexlab{a}.
\newblock \showarticletitle{Reinforcement Learning for Weakly Supervised
  Temporal Grounding of Natural Language in Untrimmed Videos}. In
  \bibinfo{booktitle}{\emph{Proceedings of the 28th ACM International
  Conference on Multimedia}} (Seattle, WA, USA). \bibinfo{pages}{1283–1291}.
\newblock


\bibitem[\protect\citeauthoryear{Wu, Li, Liu, and Lin}{Wu
  et~al\mbox{.}}{2020b}]%
        {Wu2020TreeStructuredPB}
\bibfield{author}{\bibinfo{person}{Jie Wu}, \bibinfo{person}{Guanbin Li},
  \bibinfo{person}{Si Liu}, {and} \bibinfo{person}{Liang Lin}.}
  \bibinfo{year}{2020}\natexlab{b}.
\newblock \showarticletitle{Tree-Structured Policy based Progressive
  Reinforcement Learning for Temporally Language Grounding in Video}. In
  \bibinfo{booktitle}{\emph{Proceedings of the AAAI Conference on Artificial
  Intelligence}}.
\newblock


\bibitem[\protect\citeauthoryear{Wu, Xiong, Yu, and Lin}{Wu
  et~al\mbox{.}}{2018}]%
        {wu2018unsupervised}
\bibfield{author}{\bibinfo{person}{Zhirong Wu}, \bibinfo{person}{Yuanjun
  Xiong}, \bibinfo{person}{Stella~X Yu}, {and} \bibinfo{person}{Dahua Lin}.}
  \bibinfo{year}{2018}\natexlab{}.
\newblock \showarticletitle{Unsupervised feature learning via non-parametric
  instance discrimination}. In \bibinfo{booktitle}{\emph{Proceedings of the
  IEEE Conference on Computer Vision and Pattern Recognition}}.
  \bibinfo{pages}{3733--3742}.
\newblock


\bibitem[\protect\citeauthoryear{Xu, Xiong, Chen, and Corso}{Xu
  et~al\mbox{.}}{2015}]%
        {xu2015jointly}
\bibfield{author}{\bibinfo{person}{Ran Xu}, \bibinfo{person}{Caiming Xiong},
  \bibinfo{person}{Wei Chen}, {and} \bibinfo{person}{Jason Corso}.}
  \bibinfo{year}{2015}\natexlab{}.
\newblock \showarticletitle{Jointly modeling deep video and compositional text
  to bridge vision and language in a unified framework}. In
  \bibinfo{booktitle}{\emph{Proceedings of the AAAI Conference on Artificial
  Intelligence}}, Vol.~\bibinfo{volume}{29}.
\newblock


\bibitem[\protect\citeauthoryear{Yu, Dohan, Le, Luong, Zhao, and Chen}{Yu
  et~al\mbox{.}}{2018a}]%
        {wei2018fast}
\bibfield{author}{\bibinfo{person}{Adams~Wei Yu}, \bibinfo{person}{David
  Dohan}, \bibinfo{person}{Quoc Le}, \bibinfo{person}{Thang Luong},
  \bibinfo{person}{Rui Zhao}, {and} \bibinfo{person}{Kai Chen}.}
  \bibinfo{year}{2018}\natexlab{a}.
\newblock \showarticletitle{Fast and Accurate Reading Comprehension by
  Combining Self-Attention and Convolution}. In
  \bibinfo{booktitle}{\emph{International Conference on Learning
  Representations}}.
\newblock


\bibitem[\protect\citeauthoryear{Yu, Lin, Shen, Yang, Lu, Bansal, and Berg}{Yu
  et~al\mbox{.}}{2018b}]%
        {yu2018mattnet}
\bibfield{author}{\bibinfo{person}{Licheng Yu}, \bibinfo{person}{Zhe Lin},
  \bibinfo{person}{Xiaohui Shen}, \bibinfo{person}{Jimei Yang},
  \bibinfo{person}{Xin Lu}, \bibinfo{person}{Mohit Bansal}, {and}
  \bibinfo{person}{Tamara~L Berg}.} \bibinfo{year}{2018}\natexlab{b}.
\newblock \showarticletitle{Mattnet: Modular attention network for referring
  expression comprehension}. In \bibinfo{booktitle}{\emph{Proceedings of the
  IEEE Conference on Computer Vision and Pattern Recognition}}.
  \bibinfo{pages}{1307--1315}.
\newblock


\bibitem[\protect\citeauthoryear{Yuan, Ma, Wang, Liu, and Zhu}{Yuan
  et~al\mbox{.}}{2019a}]%
        {yuan2019semantic}
\bibfield{author}{\bibinfo{person}{Yitian Yuan}, \bibinfo{person}{Lin Ma},
  \bibinfo{person}{Jingwen Wang}, \bibinfo{person}{Wei Liu}, {and}
  \bibinfo{person}{Wenwu Zhu}.} \bibinfo{year}{2019}\natexlab{a}.
\newblock \showarticletitle{Semantic Conditioned Dynamic Modulation for
  Temporal Sentence Grounding in Videos}. In \bibinfo{booktitle}{\emph{Advances
  in Neural Information Processing Systems}}. \bibinfo{pages}{536--546}.
\newblock


\bibitem[\protect\citeauthoryear{Yuan, Mei, and Zhu}{Yuan
  et~al\mbox{.}}{2019b}]%
        {Yuan2019ToFW}
\bibfield{author}{\bibinfo{person}{Yitian Yuan}, \bibinfo{person}{Tao Mei},
  {and} \bibinfo{person}{Wenwu Zhu}.} \bibinfo{year}{2019}\natexlab{b}.
\newblock \showarticletitle{To Find Where You Talk: Temporal Sentence
  Localization in Video with Attention Based Location Regression}. In
  \bibinfo{booktitle}{\emph{Proceedings of the AAAI Conference on Artificial
  Intelligence}}, Vol.~\bibinfo{volume}{33}. \bibinfo{pages}{9159--9166}.
\newblock


\bibitem[\protect\citeauthoryear{Zeng, Xu, Huang, Chen, Tan, and Gan}{Zeng
  et~al\mbox{.}}{2020}]%
        {zeng2020dense}
\bibfield{author}{\bibinfo{person}{Runhao Zeng}, \bibinfo{person}{Haoming Xu},
  \bibinfo{person}{Wenbing Huang}, \bibinfo{person}{Peihao Chen},
  \bibinfo{person}{Mingkui Tan}, {and} \bibinfo{person}{Chuang Gan}.}
  \bibinfo{year}{2020}\natexlab{}.
\newblock \showarticletitle{Dense regression network for video grounding}. In
  \bibinfo{booktitle}{\emph{Proceedings of the IEEE/CVF Conference on Computer
  Vision and Pattern Recognition}}. \bibinfo{pages}{10287--10296}.
\newblock


\bibitem[\protect\citeauthoryear{Zhang, Hu, Lee, Zhao, Chammas, Jain, Ie, and
  Sha}{Zhang et~al\mbox{.}}{2020a}]%
        {zhang2020hierarchical}
\bibfield{author}{\bibinfo{person}{Bowen Zhang}, \bibinfo{person}{Hexiang Hu},
  \bibinfo{person}{Joonseok Lee}, \bibinfo{person}{Ming Zhao},
  \bibinfo{person}{Sheide Chammas}, \bibinfo{person}{Vihan Jain},
  \bibinfo{person}{Eugene Ie}, {and} \bibinfo{person}{Fei Sha}.}
  \bibinfo{year}{2020}\natexlab{a}.
\newblock \showarticletitle{A Hierarchical Multi-Modal Encoder for Moment
  Localization in Video Corpus}.
\newblock \bibinfo{journal}{\emph{arXiv preprint arXiv:2011.09046}}
  (\bibinfo{year}{2020}).
\newblock


\bibitem[\protect\citeauthoryear{Zhang, Hu, and Sha}{Zhang
  et~al\mbox{.}}{2018}]%
        {zhang2018cross}
\bibfield{author}{\bibinfo{person}{Bowen Zhang}, \bibinfo{person}{Hexiang Hu},
  {and} \bibinfo{person}{Fei Sha}.} \bibinfo{year}{2018}\natexlab{}.
\newblock \showarticletitle{Cross-modal and hierarchical modeling of video and
  text}. In \bibinfo{booktitle}{\emph{Proceedings of the European Conference on
  Computer Vision (ECCV)}}. \bibinfo{pages}{374--390}.
\newblock


\bibitem[\protect\citeauthoryear{Zhang, Sun, Jing, Zhen, Zhou, and Goh}{Zhang
  et~al\mbox{.}}{2021}]%
        {zhang2021qa4nlvl}
\bibfield{author}{\bibinfo{person}{Hao Zhang}, \bibinfo{person}{Aixin Sun},
  \bibinfo{person}{Wei Jing}, \bibinfo{person}{Liangli Zhen},
  \bibinfo{person}{Joey~Tianyi Zhou}, {and} \bibinfo{person}{Rick Siow~Mong
  Goh}.} \bibinfo{year}{2021}\natexlab{}.
\newblock \showarticletitle{Natural Language Video Localization: A Revisit in
  Span-based Question Answering Framework}.
\newblock \bibinfo{journal}{\emph{IEEE Transactions on Pattern Analysis and
  Machine Intelligence}} (\bibinfo{year}{2021}).
\newblock


\bibitem[\protect\citeauthoryear{Zhang, Sun, Jing, and Zhou}{Zhang
  et~al\mbox{.}}{2020c}]%
        {zhang2020vslnet}
\bibfield{author}{\bibinfo{person}{Hao Zhang}, \bibinfo{person}{Aixin Sun},
  \bibinfo{person}{Wei Jing}, {and} \bibinfo{person}{Joey~Tianyi Zhou}.}
  \bibinfo{year}{2020}\natexlab{c}.
\newblock \showarticletitle{Span-based Localizing Network for Natural Language
  Video Localization}. In \bibinfo{booktitle}{\emph{Proceedings of the 58th
  Annual Meeting of the Association for Computational Linguistics}}.
  \bibinfo{publisher}{Association for Computational Linguistics},
  \bibinfo{address}{Online}, \bibinfo{pages}{6543--6554}.
\newblock


\bibitem[\protect\citeauthoryear{Zhang, Peng, Fu, and Luo}{Zhang
  et~al\mbox{.}}{2020b}]%
        {zhang2020learning}
\bibfield{author}{\bibinfo{person}{Songyang Zhang}, \bibinfo{person}{Houwen
  Peng}, \bibinfo{person}{Jianlong Fu}, {and} \bibinfo{person}{Jiebo Luo}.}
  \bibinfo{year}{2020}\natexlab{b}.
\newblock \showarticletitle{Learning 2D Temporal Adjacent Networks forMoment
  Localization with Natural Language}. In \bibinfo{booktitle}{\emph{Proceedings
  of the AAAI Conference on Artificial Intelligence}}.
\newblock


\bibitem[\protect\citeauthoryear{Zhang, Su, and Luo}{Zhang
  et~al\mbox{.}}{2019b}]%
        {zhang2019exploiting}
\bibfield{author}{\bibinfo{person}{Songyang Zhang}, \bibinfo{person}{Jinsong
  Su}, {and} \bibinfo{person}{Jiebo Luo}.} \bibinfo{year}{2019}\natexlab{b}.
\newblock \showarticletitle{Exploiting Temporal Relationships in Video Moment
  Localization with Natural Language}. In \bibinfo{booktitle}{\emph{Proceedings
  of the 27th ACM International Conference on Multimedia}} (Nice, France).
  \bibinfo{pages}{1230–1238}.
\newblock


\bibitem[\protect\citeauthoryear{Zhang, Lin, Zhao, and Xiao}{Zhang
  et~al\mbox{.}}{2019a}]%
        {zhu2019cross}
\bibfield{author}{\bibinfo{person}{Zhu Zhang}, \bibinfo{person}{Zhijie Lin},
  \bibinfo{person}{Zhou Zhao}, {and} \bibinfo{person}{Zhenxin Xiao}.}
  \bibinfo{year}{2019}\natexlab{a}.
\newblock \showarticletitle{Cross-Modal Interaction Networks for Query-Based
  Moment Retrieval in Videos}. In \bibinfo{booktitle}{\emph{Proceedings of the
  42nd International ACM SIGIR Conference on Research and Development in
  Information Retrieval}}. \bibinfo{pages}{655–664}.
\newblock


\bibitem[\protect\citeauthoryear{Zhu and Yang}{Zhu and Yang}{2020}]%
        {zhu2020actbert}
\bibfield{author}{\bibinfo{person}{Linchao Zhu} {and} \bibinfo{person}{Yi
  Yang}.} \bibinfo{year}{2020}\natexlab{}.
\newblock \showarticletitle{ActBERT: Learning Global-Local Video-Text
  Representations}. In \bibinfo{booktitle}{\emph{Proceedings of the IEEE/CVF
  Conference on Computer Vision and Pattern Recognition}}.
  \bibinfo{pages}{8746--8755}.
\newblock


\bibitem[\protect\citeauthoryear{Zhuang, Zhai, and Yamins}{Zhuang
  et~al\mbox{.}}{2019}]%
        {zhuang2019local}
\bibfield{author}{\bibinfo{person}{Chengxu Zhuang}, \bibinfo{person}{Alex~Lin
  Zhai}, {and} \bibinfo{person}{Daniel Yamins}.}
  \bibinfo{year}{2019}\natexlab{}.
\newblock \showarticletitle{Local aggregation for unsupervised learning of
  visual embeddings}. In \bibinfo{booktitle}{\emph{Proceedings of the IEEE/CVF
  International Conference on Computer Vision}}. \bibinfo{pages}{6002--6012}.
\newblock


\end{thebibliography}

\end{document}